\documentclass[letterpaper, 10 pt, conference]{ieeeconf}

\IEEEoverridecommandlockouts                              

\usepackage{amsmath,amsfonts}
\usepackage{bm}
\usepackage{array}
\usepackage{textcomp}
\usepackage{stfloats}
\usepackage{url}
\usepackage{verbatim}
\usepackage{graphicx}
\usepackage{cite}
\usepackage{multirow}
\usepackage{booktabs}
\usepackage{color}
\usepackage{lineno}
\usepackage{subcaption}
\usepackage{algorithmic}
\usepackage[ruled]{algorithm2e}
\usepackage{hyperref}

\title{\LARGE \bf
CTSAC: Curriculum-Based Transformer Soft Actor-Critic\\ for Goal-Oriented Robot Exploration}

\author{Chunyu Yang$^{*}$, Shengben Bi$^{*}$, Yihui Xu, Xin Zhang%
\thanks{
    $^{\dagger}$ The authors are with the School of Information and Control Engineering, China University of Mining and Technology, Xuzhou, 221116, China. Correspondence to: Chunyuyang@cumt.edu.cn (Chunyu Yang). This work was supported by the National Natural Science Foundation of China under Grant 62273350. Code and videos can be found at https://github.com/ShengbenBi/CTSAC.}%
\thanks{$^{*}$ These authors are co-first authors.}
}
\begin{document}
\maketitle
\thispagestyle{empty}
\pagestyle{empty}

\begin{abstract}
With the increasing demand for efficient and flexible robotic exploration solutions, Reinforcement Learning (RL) is becoming a promising approach in the field of autonomous robotic exploration.
However, current RL-based exploration algorithms often face limited environmental reasoning capabilities, slow convergence rates, and substantial challenges in Sim-To-Real (S2R) transfer.
To address these issues, we propose a Curriculum Learning-based Transformer Reinforcement Learning Algorithm (CTSAC) aimed at improving both exploration efficiency and transfer performance.
To enhance the robot's reasoning ability, a Transformer is integrated into the perception network of the Soft Actor-Critic (SAC) framework, leveraging historical information to improve the farsightedness of the strategy.
A periodic review-based curriculum learning is proposed, which enhances training efficiency while mitigating catastrophic forgetting during curriculum transitions.
Training is conducted on the ROS-Gazebo continuous robotic simulation platform, with LiDAR clustering optimization to further reduce the S2R gap.
Experimental results demonstrate the CTSAC algorithm outperforms the state-of-the-art non-learning and learning-based algorithms in terms of success rate and success rate-weighted exploration time.
Moreover, real-world experiments validate the strong S2R transfer capabilities of CTSAC.
%%Code and videos can be found at https://github.com/ShengbenBi/CTSAC.
\end{abstract}

%%%%%%%%%%%%%%%%%%%%%%%%%%%%%%%%%%%%%%%%%%%%%%%%%%%%%%%%%%%%%%%%%%%%%%%%%%%%%%%%
\section{INTRODUCTION}
Autonomous Exploration (AE) enables robots to perceive their environment, plan paths, reach goals, or build maps while avoiding obstacles. This capability is crucial in fields such as space exploration, search and rescue, and reconnaissance. At each time step, the robot must plan the shortest and most efficient path based on the currently observed data. This task falls under sequential decision-making and is classified as an NP-Hard\cite{c1}.

Currently, AE is classified into non-learning-based and learning-based approaches, depending on the methods employed\cite{c2}. Non-learning-based methods primarily include information-theoretic\cite{c3,c4}, frontier-based\cite{c5}, and random sampling-based approaches\cite{c1,c6,c7,c8}. Human-designed exploration algorithms often rely on strong assumptions about the environment and the task. However, due to their limited reasoning ability regarding the environment\cite{c9,c10}, as the environment is gradually revealed, the optimal paths generated from partially known maps frequently become suboptimal \cite{c11}, leading to inefficiencies in traditional exploration algorithms.

\begin{figure}[htbp]
        \centerline{\includegraphics[width=0.4\textwidth]{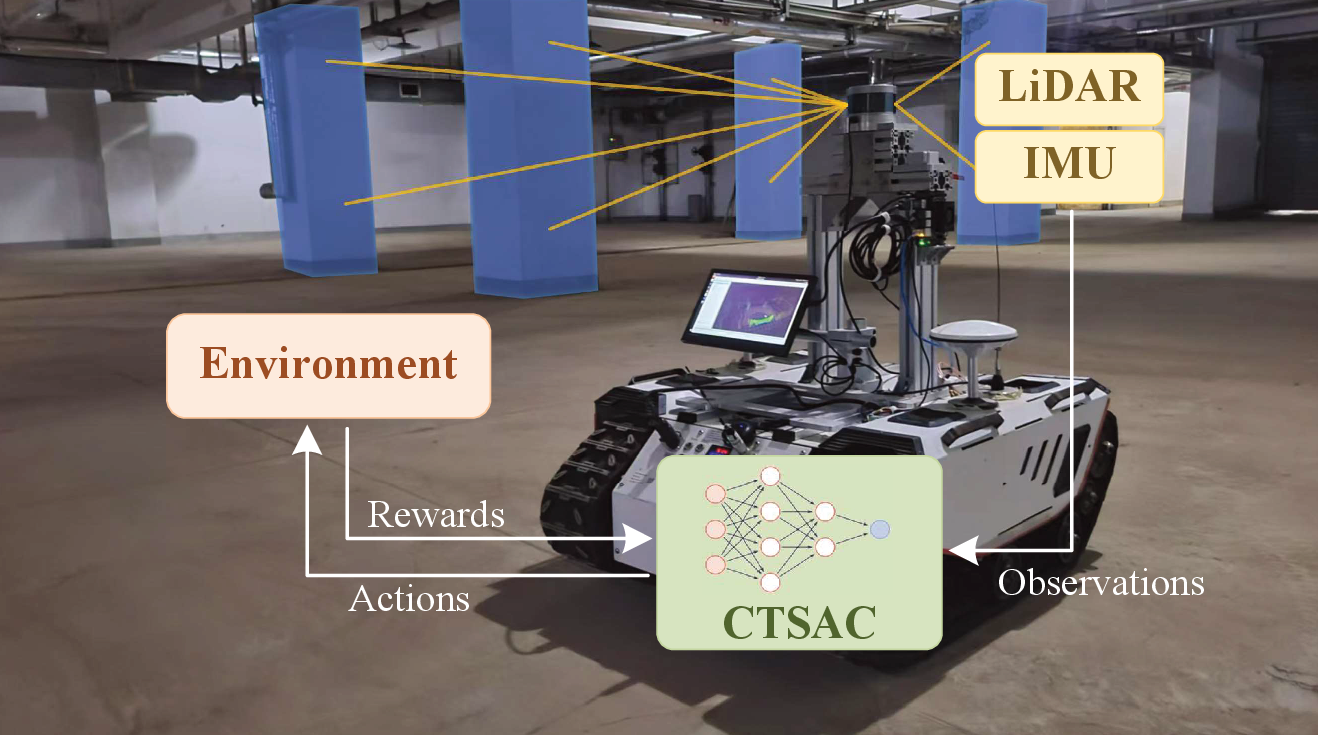}}
        \caption{\textbf{CTSAC-based autonomous robot exploration system.} The robot perceives its environment using LiDAR and IMU sensors, and makes decisions through the CTSAC algorithm, enabling efficient exploration the environment.}
        \label{fig1}
\end{figure}

Reinforcement Learning (RL) offers a promising alternative, allowing agents to autonomously develop exploration strategies through their interactions with the environment. However, RL by itself often falls short in addressing the complex decision-making, To overcome this limitation, \cite{c12} proposed a method that uses Convolutional Neural Networks (CNN) to extract information from raw sensor data and employs Deep Q-Networks (DQN) to generate robot motion commands. Research by \cite{c9,c13,c14,c15,c16,c17} focused on enhancing prediction accuracy and improving the long-term foresight of agent by innovating network architectures. However, all of these methods rely solely on current observations for decision-making, overlooking the importance of historical information. During experiments, it was found that the robot is prone to get stuck in a local optimum and enter a wandering state, making it difficult to escape from the local optimum. As a typical example of Partially Observable Markov Decision Processes (POMDPs) \cite{c18}, AE highlights the limitations of relying solely on current observations for accurate decision-making. Consequently, incorporating historical observations is essential for selecting the optimal action \cite{c19}.

To enable RL to generalize effectively, it is typically trained in complex environments to cover a broad range of scenarios. However, training directly in such environments introduces the risk of model divergence and instability \cite{c20,c21,c22}. Reference \cite{c23} has shown that the Soft Actor-Critic (SAC) algorithm \cite{c24} can learn better exploration strategies, but the training time for SAC can extend up to 8.5 days. Additionally, RL often depends on large-scale CNNs to process LiDAR and image data \cite{c9,c25,c26,c27}, which exacerbates challenges related to training stability and convergence speed. To address these issues, \cite{c29,c25,c28} propose the use of curriculum learning to accelerate training, while \cite{c28} introduces the concept of a cumulative course of study to address catastrophic forgetting \cite{c30}. Despite these advancements, applying these methods to continuous spaces remains a significant challenge \cite{c32,c31}. Consequently, most RL-based methods are still trained in table-based simulation environments \cite{c9,c25,c26,c33}, which fail to account for the robot's movement, the authenticity of the environment, and other physical factors, making it difficult to apply these approaches to real-world scenarios\cite{c34}.

We integrate the powerful sequential perception capabilities of Transformers \cite{c35} into the SAC framework to utilize both the robot's current and historical state information. This integration enables the agent to uncover latent relationships between different regions of the environment and its past states, helping guide the robot out of looping states when trapped in a local optimum, thus enhancing its decision-making and reasoning abilities. Additionally, we propose a periodic review-based curriculum learning strategy in the continuous environment of the Gazebo-ROS simulation. This strategy gradually increases task difficulty while revisiting previous tasks to retain knowledge and mitigate catastrophic forgetting. Leveraging the continuous nature of the environment, this approach improves the Sim-To-Real (S2R) transferability of the trained model. Furthermore, building upon the work outlined in \cite{c32}, we have optimized LiDAR clustering based on the robot's direction to enhance the segmentation method. This optimization improves the robot's perception of its environment while reducing the dimensionality of the original LiDAR data, further narrowing the S2R gap and facilitating the model's deployment in real-world environments, as shown in Fig. 1.

%%我们在sac中引入了强大序列感知能力的Transformer用于推理机器人的当前帧和历史状态信息，使代理能够揭示环境不同区域之间的潜在关系和机器人的历史状态信息，以增强智能体的决策和推理能力以增强智能体的决策和推理能力，帮助机器人摆脱局部最优状态。
%%我们将机器人的历史状态信息考虑在内，将Transformer的强大的序列感知能力与Soft Actor-Critic（SAC）基于熵的强化学习算法相结合，使代理能够揭示环境不同区域之间的潜在关系和机器人的历史状态信息，以增强智能体的决策和推理能力，helping guide the robot out of looping states when it becomes trapped in a local optimum.

%%Additionally, we propose a periodic review-based curriculum learning strategy, which gradually increases task difficulty while revisiting previous environments to retain knowledge and mitigate catastrophic forgetting.
%%Additionally, we propose a periodic review-based curriculum learning strategy在连续空间的gazebo-ros仿真环境中, which gradually increases task difficulty while revisiting previous 课程 to retain knowledge and mitigate catastrophic forgetting，通过连续的仿真环境，提高了训练出的模型的sim-to-real（S2R）转移能力。此外，我们在\cite{c32}激光雷达预处理的基础上，按照机器人的方向优化了optimize the segmentation method，使机器人在前进时对前方的感知更加详细的同时降低了原始激光雷达的维度，通过预处理进一步降低了sim-to-real（S2R）的gap，making it easier to apply the model to the actual environment as in Fig.1

%%leveraging dimensionality-reduced LiDAR data to bridge the sim-to-real (S2R) gap provides a foundation for the deployment of RL-based systems in real-world environments\cite{c32}.making it easier to apply the model to the actual environment as in Fig.1

The novelties and contributions of this work are: 

\begin{itemize}
        \item A Transformer-based reinforcement learning algorithm is proposed to reason with both the robot's historical state information and the environmental context, addressing the issue of the robot's lack of long-term vision and its tendency to get stuck in loops.
        \item A periodic review-based curriculum learning strategy is introduced to improve the efficiency and stability of the training process, addressing the problem of catastrophic forgetting.
        \item A LiDAR clustering optimization regarding the robot's direction is used to reduce the sim-to-real gap. A ROS-Gazebo-Pytorch continuous space training platform is employed, and the proposed algorithm has been validated in both simulated and real-world scenarios.
        \end{itemize}

\section{PROBLEM FORMULATION}

AE is essentially a type of sequential decision-making problem. Since exploration takes place in an unknown environment, the robot cannot observe the full global state and can only acquire partial information through sensors. Accordingly, AE can be described as the following Markov Decision Process: Let ${s_t}$ denote the state of the robot at time $t$, which comprises information about surrounding obstacles obtained from the preprocessed LiDAR observations ${l_1}, ..., {l_d}$, the robot's linear velocity ${v_r}$, angular velocity $\omega_r$, as well as the relative distance ${d_t}$ and angle ${\theta_t}$ between the goal point and the robot. Based on the policy $\pi $, the robot selects an action ${a_t}$, which comprises a linear velocity ${v_c}$ and an angular velocity $\omega_c$, within its kinematic limits. Subsequently, the environment provides a reward ${r_t}$ for the robot's action, leading to a state transition to the next state ${s_{t + 1}}$. This process repeats until the robot reaches the preset target state. 

\section{CTSAC APPROACH}
\subsection{End-to-end goal-oriented robot exploration framework}
Fig. 2 illustrates the end-to-end AE system architecture, referred to as CTSAC, where the robot, equipped with a LiDAR and an Inertial Measurement Unit (IMU), performs environmental perception and state estimation.

\begin{figure*}[htbp]
    \vskip 0.18cm
    \centerline{\includegraphics[width=0.85\textwidth]{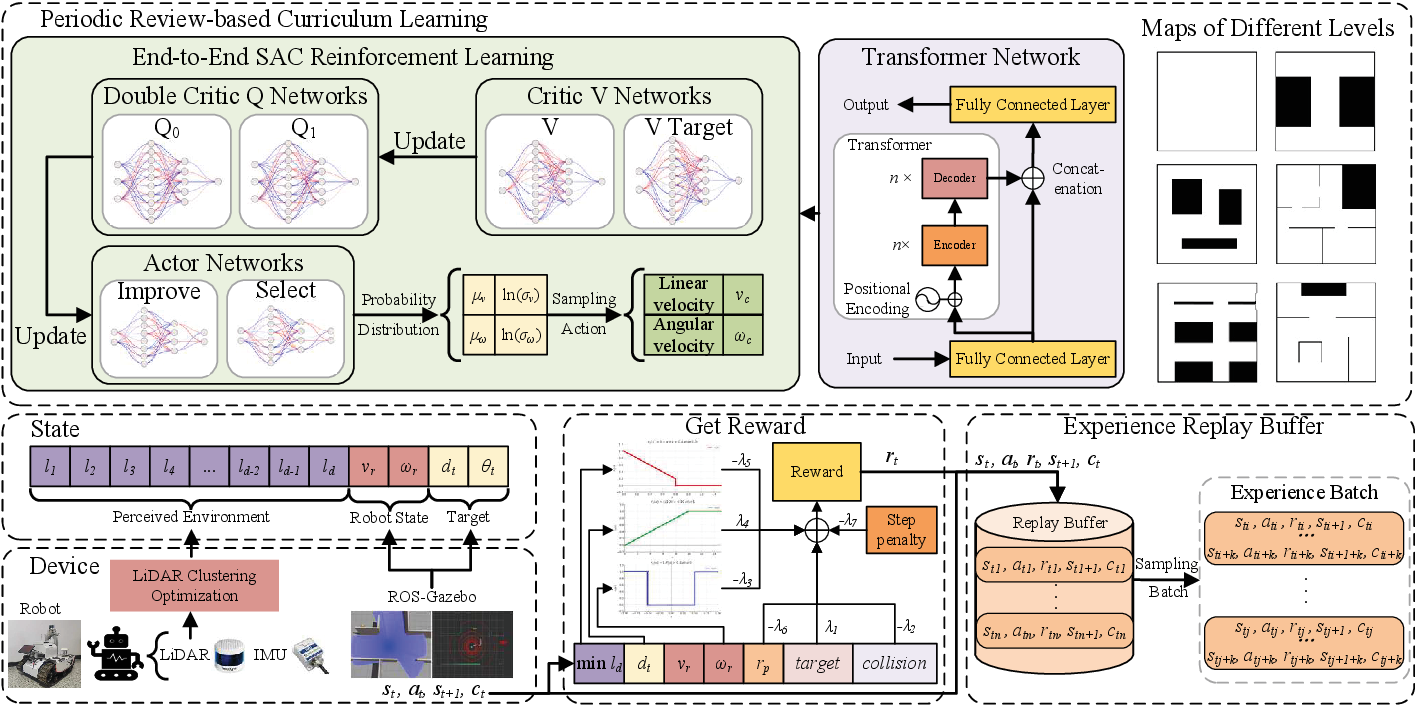}}
    \caption{Overview of the CTSAC autonomous exploration system architecture}
    \label{fig2}
\end{figure*}
First, the LiDAR data undergoes preprocessing, where the scanning area is divided into multiple segments based on its dimensional properties, and the LiDAR data within each segment is clustered. Typically, the area is evenly partitioned into several parts\cite{c32,c31}. However, for a robot moving forward, perception in the forward direction is more critical. Therefore, we optimize the segmentation method regarding robot direction in Fig. 3. Let $d$ denote the segmentation dimension of the LiDAR, with $\Delta {\theta _m}$ representing the angular span of the $m$-th segment:
\begin{equation}
    \Delta\theta_m = \begin{cases}
        4\pi \ / \ (3d-4) & \text{if } m < 3d \ / \ 4 \\
        4\pi(3d - 8) \ / \ d(3d - 4) & \text{if } m \ge 3d \ / \ 4
    \end{cases}
\end{equation}

\begin{figure}[htbp]
    \centerline{\includegraphics[width=0.4\textwidth]{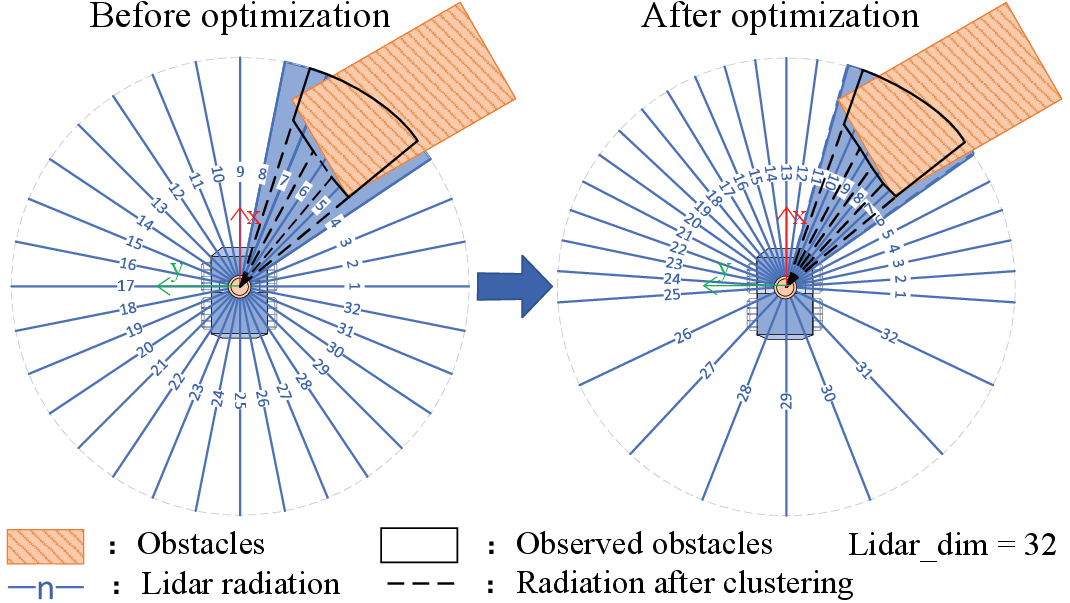}}
    \caption{\textbf{LiDAR clustering optimization regarding robot direction.} The figure illustrates an example with $d$ = 32.}
    \label{fig3}
\end{figure}
With the same LiDAR dimension $d$, the optimized segmentation method enhances the robot's forward perception for identical obstacles and reduces the network's decision-making time to approximately 3.7 $ms$. 

Besides, the reward setting during the training process is also particularly important. The reward function is mainly composed of seven parts:

\begin{equation}
    R = \left\{
    \begin{array}{ll}
        \lambda_1 & \text{if goal} \\
        -\lambda_2 & \text{if collision} \\
      \left.
      \begin{array}{l}
        \hspace{-0.67em}-\lambda_3 \cdot r_1(\omega_r) + \lambda_4 \cdot r_2(d_t) \\
        \hspace{-0.67em}-\lambda_5 \cdot r_3(\min l_d) - \lambda_6 \cdot r_p - \lambda_7
      \end{array}
      \right\} & \text{otherwise}
    \end{array}
    \right.
\end{equation}

When the robot reaches the goal, a positive reward ${\lambda _1}$ is given, and the episode is terminated. When the robot collides, a negative reward ${\lambda _2}$ is given, and the episode is similarly terminated.
\subsubsection{Turning penalty}
A penalty mechanism is implemented to limit the robot's turning frequency. When the angular velocity $\omega_r$ exceeds 0.5 $\text{rad/s}$, a penalty of 1 is applied.

\begin{equation}
r_1(\omega_r)=\begin{cases}1&\text{if}\mid \omega_r\mid>0.5\\0&\text{otherwise}\end{cases}
\end{equation}

\subsubsection{Goal proximity reward}
To encourage the robot to move towards the goal, a reward is assigned when the goal distance ${d_t}$ is less than 10 $\mathrm{m}$. To prevent excessive rewards from distorting the overall reward structure, the coefficient ${\lambda _4}$ is set relatively small.

\begin{equation}
    r_2(d_t) = \begin{cases}
        d_t \ / \ 10 & \text{if } d_t < 10 \\
        1 & \text{otherwise}
    \end{cases}
\end{equation}

\subsubsection{Obstacle proximity penalty}

To ensure that the robot maintains a safe distance from obstacles, we process the LiDAR data to determine the minimum distance, denoted as $\min {l_d}$, to any obstacles. When $\min {l_d}$ is less than 1 $\mathrm{m}$, a certain penalty is applied.

\begin{equation}
r_3(\min l_d)=\begin{cases}1-\min l_d&\text{if } \min l_d<1\\0&\text{otherwise}\end{cases}
\end{equation}

\subsubsection{Wandering penalty}
To prevent the robot from falling into a local optimum (e.g., spinning or wandering in one location) and to encourage broader exploration, we introduce a wandering penalty mechanism. This mechanism counts the number of times the Manhattan distance between the current position $(x,y)$ and the stored historical positions $(x_i,y_i)$ within the episode falls below a threshold $\delta $, which is used as the penalty factor ${R_p}$, ${\bf{1}}( \cdot )$ is an indicator function that returns 1 if the condition is met and 0 otherwise.
\begin{equation}
    d_m(x,y,x_{i},y_{i}) = \mid x - x_{i} \mid + \mid y - y_{i} \mid
\end{equation}
\begin{equation}
    r_p = \sum_{i=1}^n \mathbf{1}(d_m(x,y,x_i,y_i) < \delta)
\end{equation}
\subsubsection{Step penalty}
To encourage the robot to reach the goal as quickly as possible and minimize unnecessary detours, a constant step penalty ${\lambda _7}$ is applied at each step. This incentivizes the robot to follow a more direct path, effectively reducing inefficient exploration.

To facilitate S2R transfer, we utilized the continuous-space Gazebo simulation for training and built the RL model with PyTorch, integrating ROS for data transmission into a unified ROS-Gazebo-PyTorch training platform. To enhance realism in the Gazebo simulation, we introduced environmental noise, including LiDAR sensor noise and data transmission delays, etc. To improve the robot’s sampling efficiency, a diverse range of randomized scenarios was incorporated, including random obstacles generation as well as randomized initial positions and goal locations for the robot. To mitigate the slow simulation speed in Gazebo, we optimized the simulation rate parameters, achieving an approximately tenfold increase in simulation speed. The CTSAC algorithm was trained for approximately 90,000 steps, reducing the total training time to around 15.2 $h$.
%%ROS is used as the data transmission framework, establishing a ROS-Gazebo-PyTorch integrated simulation training platform. In the Gazebo simulation environment, we introduced environmental noise such as LiDAR sensors and data transmission delays to fully simulate real-world scenarios. To improve the robot's sampling efficiency, a large number of random scenarios were introduced during the environment setup, including the random generation of obstacles, and random settings for the robot's initial position and goal location. Additionally, to address the slow simulation speed in Gazebo, we increased the simulation speed by about 10 times by adjusting the simulation rate parameters. During training, the CTSAC algorithm was trained in 90000 steps and the time was reduced to 15.2h.

\subsection{Transformer-based Soft Actor-Critic}
\begin{figure}[htbp]
    \vspace{0.2cm} 
    \centerline{\includegraphics[width=0.47\textwidth]{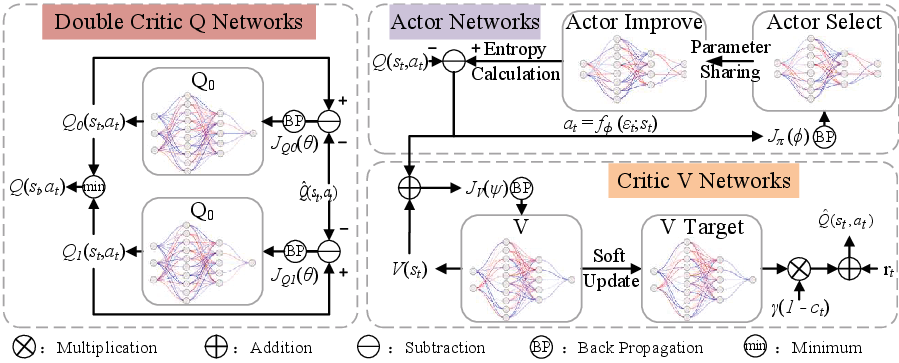}}
    \caption{SAC architecture diagram.}
    \label{fig4}
\end{figure}
SAC is an off-policy actor-critic framework based on maximum entropy reinforcement learning, as shown in Fig. 4. Its goal is to maximize the expected reward while also maximizing the entropy of the policy, ensuring the successful completion of the task while promoting as much exploration as possible. SAC consists of three types of neural networks: the Actor networks, double critic Q networks, and critic V networks.

% then the optimal strategy is:$\pi^*=\arg\max_\pi\mathbb{E}_{\tau\sim\pi}\left[\sum_{t=0}^\infty\gamma^t\left(r(\mathbf{s}_t,\mathbf{a}_t)+\alpha\mathcal{H}(\pi(\cdot|\mathbf{s}_t))\right)\right]$

% The objective function of the Actor network is:$J_\pi(\phi)=\mathbb{E}_{\mathbf{s}_t \sim \mathcal{D}, \epsilon_t \sim \mathcal{N}}\left[\alpha \log \pi_\phi(a_t|s_t) - Q_\theta(\mathbf{s}_t, a_t)\right]$

% The objective function of the V network is:$J_V(\psi)=\mathbb{E}_{\mathbf{s}_t\sim\mathcal{D}}\left[\frac12{\left(V_\psi(\mathbf{s}_t)-\mathbb{E}_{\mathbf{a}_t\sim\pi_\phi}{\left[Q_\theta(\mathbf{s}_t,\mathbf{a}_t)-\alpha\log\pi_\phi(a_t|s_t)\right]}\right)}^2\right]$

% The objective function of the Q network is:$J_Q(\theta)=\mathbb{E}_{(\mathbf{s}_t,\mathbf{a}_t)\sim D}\left[\frac{1}{2}\Big(Q_\theta(\mathbf{s}_t,\mathbf{a}_t)-\hat{Q}(\mathbf{s}_t,\mathbf{a}_t)\Big)^2\right]$

% Among them target Q is: $\hat{Q}(\mathbf{s}_t,\mathbf{a}_t)=r(\mathbf{s}_t,\mathbf{a}_t)+\gamma(1-c_t)V_{\overline{\psi}}(\mathbf{s}_{t+1})$

The temperature coefficient $\alpha $ is a dynamic adjustment parameter of the initial temperature coefficient ${\alpha _0}$, which is set to 1 by default. $\tau $ is the temperature decay rate, set to $1 \times 10^{-6}$. During training, the number of episodes $n_e$ completed in a specific curriculum stage affects the adjustment of the temperature coefficient. When the training curriculum switches to a different stage, $n_e$ is reset to 0. At the beginning of a new environment, the agent focuses more on exploration, and as training progresses, the agent gradually shifts towards exploiting the knowledge it has acquired.
\begin{equation}
    \alpha = \alpha_0 \ / \ (1.0 + \tau \cdot n_e)
\end{equation}

In this work, Transformer is used as the core neural network structure in the SAC model, replacing the fully connected layers typically found in traditional deep reinforcement learning. The designed neural network structure is shown in Fig. 5:
\begin{figure}[!h]
    \vskip 0.18cm
    \centering
    \setcounter{subfigure}{0}
    \subfloat[Actor Select network architecture]{
        \includegraphics[width=0.95\linewidth]{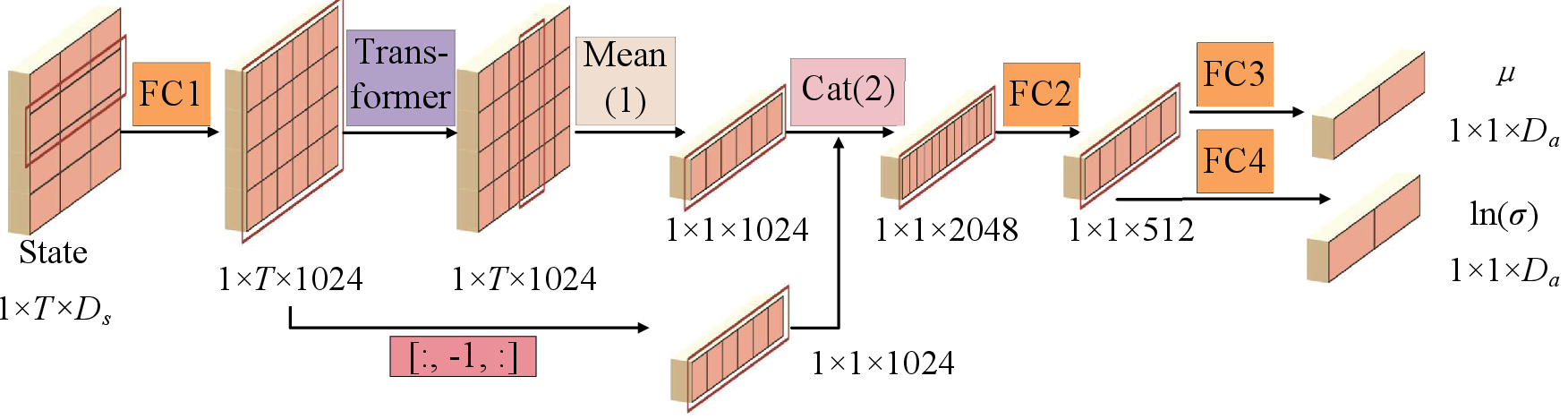}
    }
    \vspace{0.1cm}
    \subfloat[Actor Improve network architecture]{
        \includegraphics[width=0.95\linewidth]{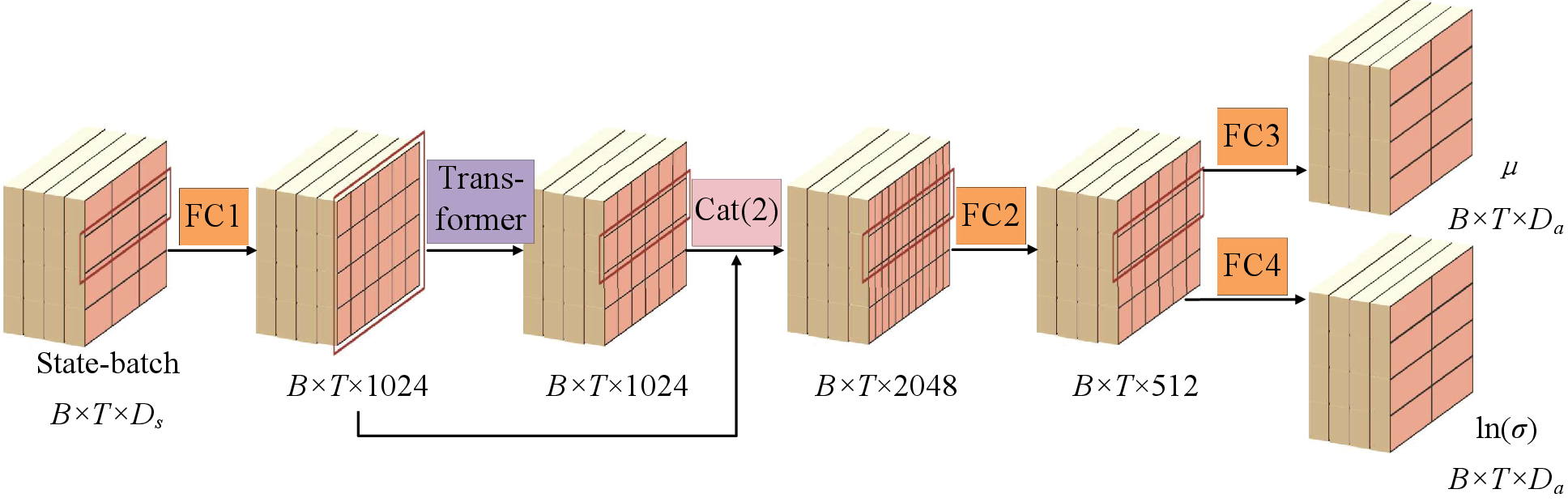}
    }
    \vspace{0.1cm}
    \subfloat[V network architecture]{
        \includegraphics[width=0.95\linewidth]{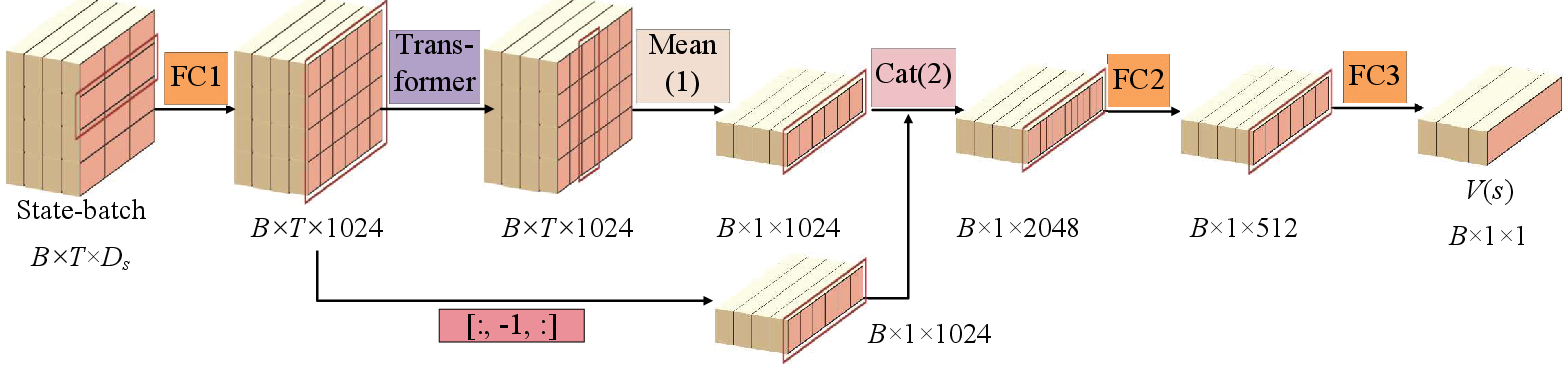}
    }
    \vspace{0.1cm}
    \subfloat[Q network architecture]{
        \includegraphics[width=0.95\linewidth]{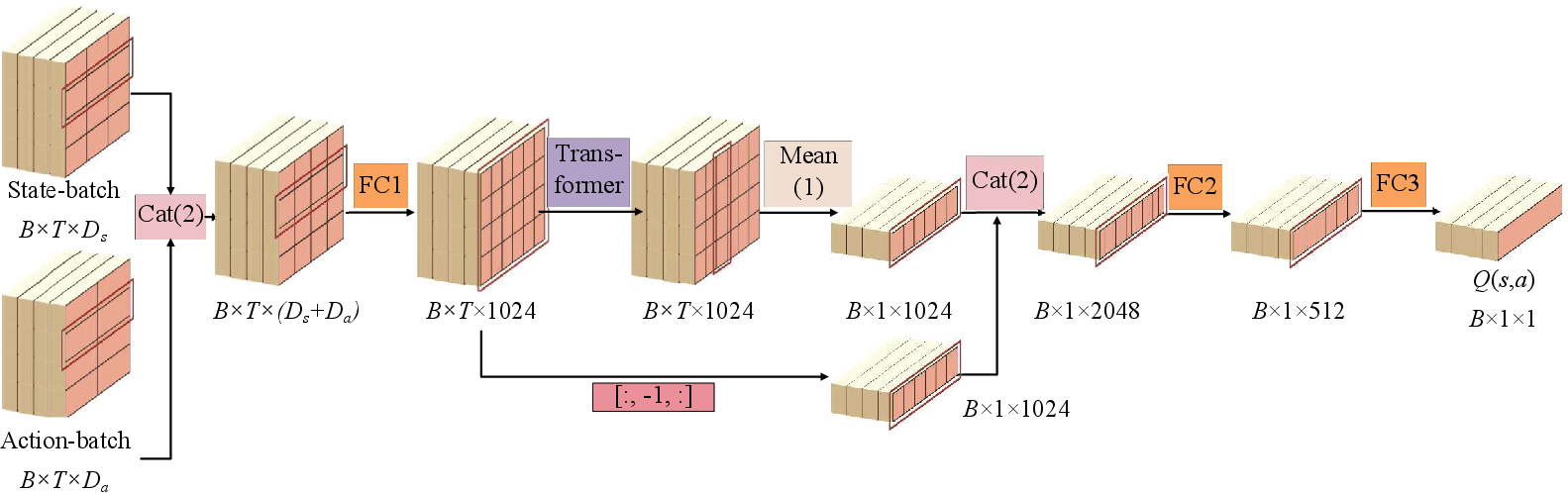}
    }
    \caption{\textbf{Transformer-based architecture for SAC network.} The $FC$ denotes a fully connected layer, while $\text{Mean}(n)$ and $\text{Cat}(n)$ represent dimension-wise averaging and concatenation operations along $n$ dimensions, respectively. The batch parameters include $B$ (batch size) and $T$ (sequence length per sample). For feature dimensions,  $D_s$ and $D_a$ correspond to the state feature dimension and action feature dimension, respectively.}
    \label{fig5}
\end{figure}

\subsubsection{Actor networks}

The Actor network is divided into two parts: Actor Select and Actor Improve, both of which share parameters. The Actor Select network is invoked during the decision-making process, where its input consists of the robot's states over $T$ steps in the environment. After dimensional expansion via a fully connected layer, the input is passed through a Transformer layer, which generates correlation information between the steps and merges it with the current step’s data. An average operation is then applied along the sequence length dimension, and the result is concatenated with the current step's information along the feature dimension. Finally, the concatenated output is passed through two fully connected layers, which generate the mean $\mu$ and the log standard deviation $\ln(\sigma)$ of the action.

% the input is passed into the Transformer layer, which generates correlation information between the steps and merges it with the current step's information. Then, the output of the Transformer layer is concatenated with an additional tensor before being passed into two fully connected layers. These layers are used to generate the mean $\mu$ and the log standard deviation $\text{log_std}$, respectively.
The Actor Improve network is invoked during the policy improvement process, receiving experience sequences from the experience replay buffer. To generate sequence-based action data for the Q network to learn, the sequence data processed by the Transformer layer is concatenated with the original data, preserving the original state information while inferring contextual insights.

% The Actor network is divided into two parts: policy selection and policy improvement, with both parts sharing parameters. In the policy selection phase, the network's input consists of all the states of the robot over $n$ steps in the environment. After being upsampled through a fully connected layer, the input is passed into the Transformer layer, which generates correlation information between the steps and merges it with the current step's information. Then, two fully connected layers are used to generate the mean and variance of the actions, respectively. 

% In the policy improvement phase, the network receives experience sequences from the experience replay buffer. The sequence data processed by the Transformer layer does not require downsampling but is directly concatenated with the upsampled original data. This design aims to generate action data with sequential dimensions for the Q network to learn.

\subsubsection{Critic V networks}
The V network consists of two components: one for estimating the current state value $V(s_t)$ and another for calculating the target value $V({{s_{t + 1}}}) $. The parameters of the V Target network are updated through a soft update from the V network. The structure of the V networks is similar to the Actor Select network, but the difference is that the V networks output only a single one-dimensional value $V(s)$ at the end.
\subsubsection{Double critic Q networks} 
The Q network consists of two Q networks with identical structures but independent parameters, which are updated separately. Ultimately, the final $Q$ value is the minimum of the two network outputs. The Q network architecture is similar to the V network, with the main difference being that the Q network takes the concatenation of the state and action batches as input, while the rest of the structure remains the same.

% During training, the neural network is updated once every two iterations. The learning rate is set to 5e-4, the discount factor $\gamma$ is set to 0.98, the experience buffer size is set to 1e5, the batch size is 256, and the random seed is set to 1. The Adam optimizer is used to optimize the neural network parameters. Additionally, to prevent model overfitting, dropout and weight decay techniques are introduced in the Transformer structure. The embedding dimension of the Transformer layer is set to 1024, with 2 encoder blocks and 2 decoder blocks, each containing 8 attention heads.

During training, the neural network is updated once every two iterations. The learning rate is set to $5 \times 10^{-4}$, the discount factor $\gamma$ is set to 0.98, the experience buffer size is set to $1 \times 10^{5}$, the batch size is 256, and the random seed is set to 1. The Adam optimizer is used to optimize the neural network parameters. Additionally, to prevent overfitting, dropout and weight decay techniques are applied within the Transformer structure. The embedding dimension of the Transformer layer is set to 1024, with 2 encoder blocks and 2 decoder blocks, each containing 8 attention heads.
\subsection{Periodic review-based curriculum learning}

To accelerate the training speed and stability of the agent and to prevent it from falling into local optima that could lead to training divergence, we employ a periodic review-based curriculum learning approach to train Transformer-based soft actor-critic (TSAC), while also mitigating the common issue of catastrophic forgetting in curriculum learning. In each training phase, revisiting previous tasks at specific intervals to ensure successful learning in new environments while effectively retaining the knowledge acquired from earlier stages.

\begin{figure}[!h]
    \centering
    \setcounter{subfigure}{0}

    % 第一行子图
    \begin{minipage}{\linewidth}
        \centering
        \subfloat{
            \includegraphics[width=0.165\linewidth, height=0.165\linewidth]{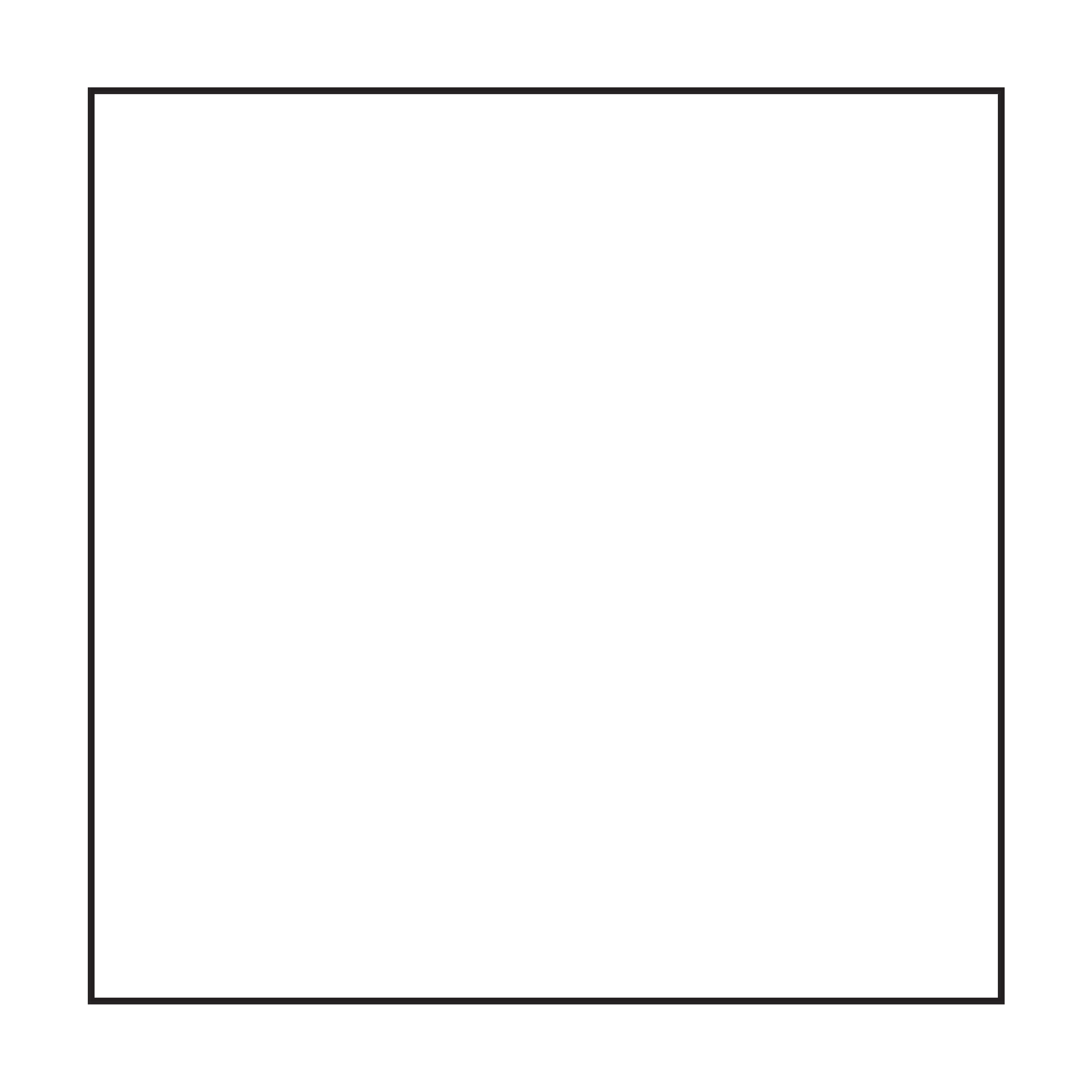}
        }\hspace{-1.2em}
        \subfloat{
            \includegraphics[width=0.165\linewidth, height=0.165\linewidth]{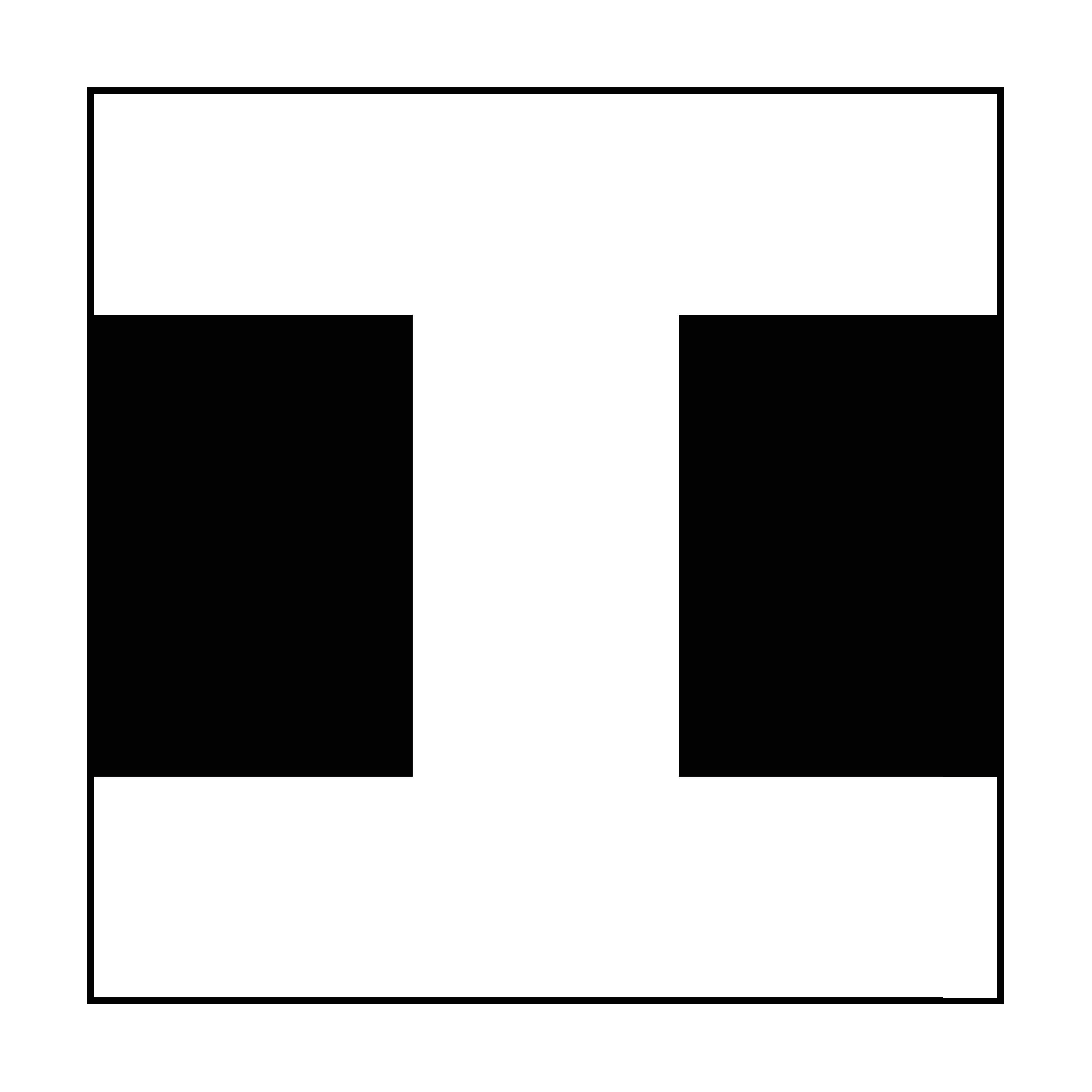}
        }\hspace{-1.2em}
        \subfloat{
            \includegraphics[width=0.165\linewidth, height=0.165\linewidth]{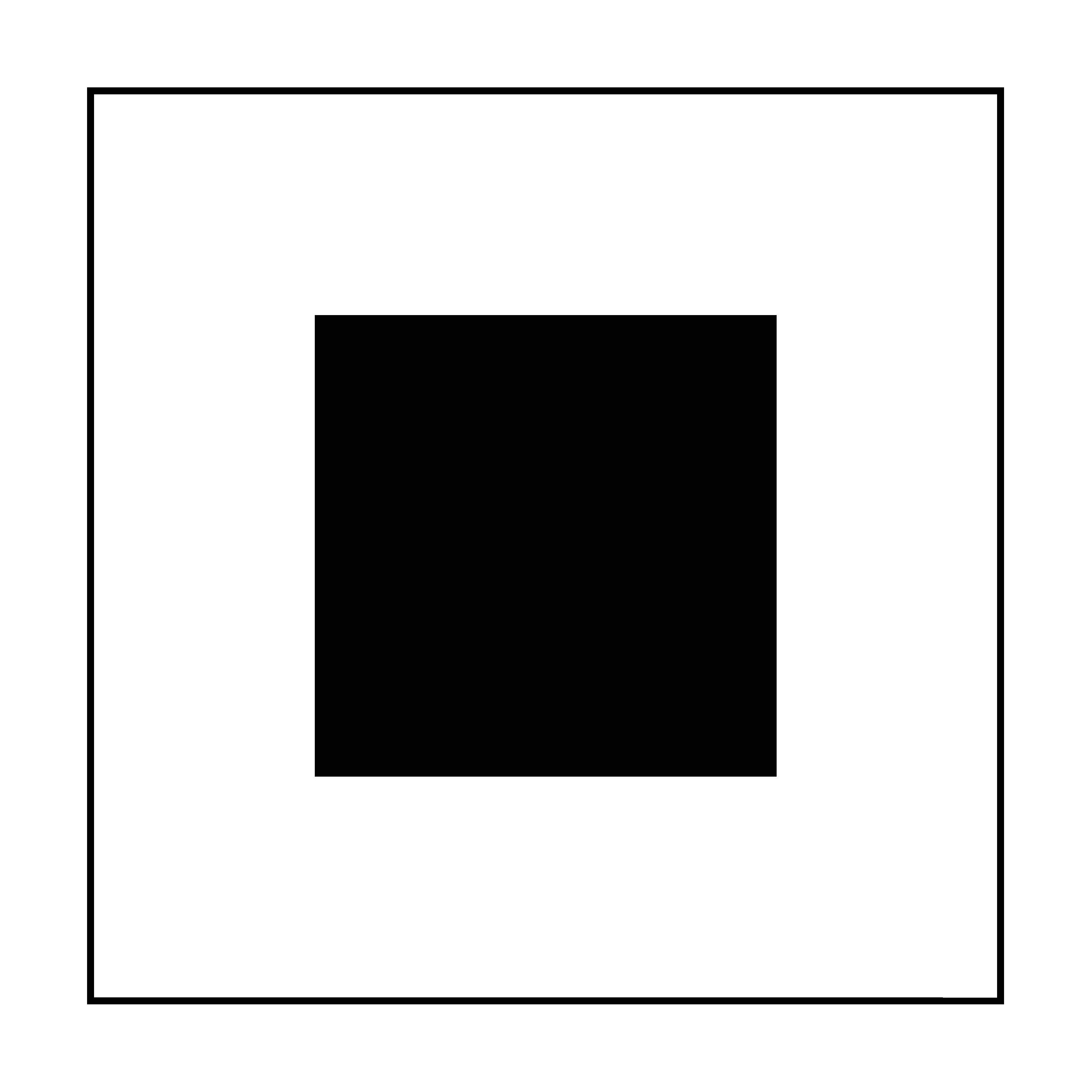}
        }\hspace{-1.2em}
        \subfloat{
            \includegraphics[width=0.165\linewidth, height=0.165\linewidth]{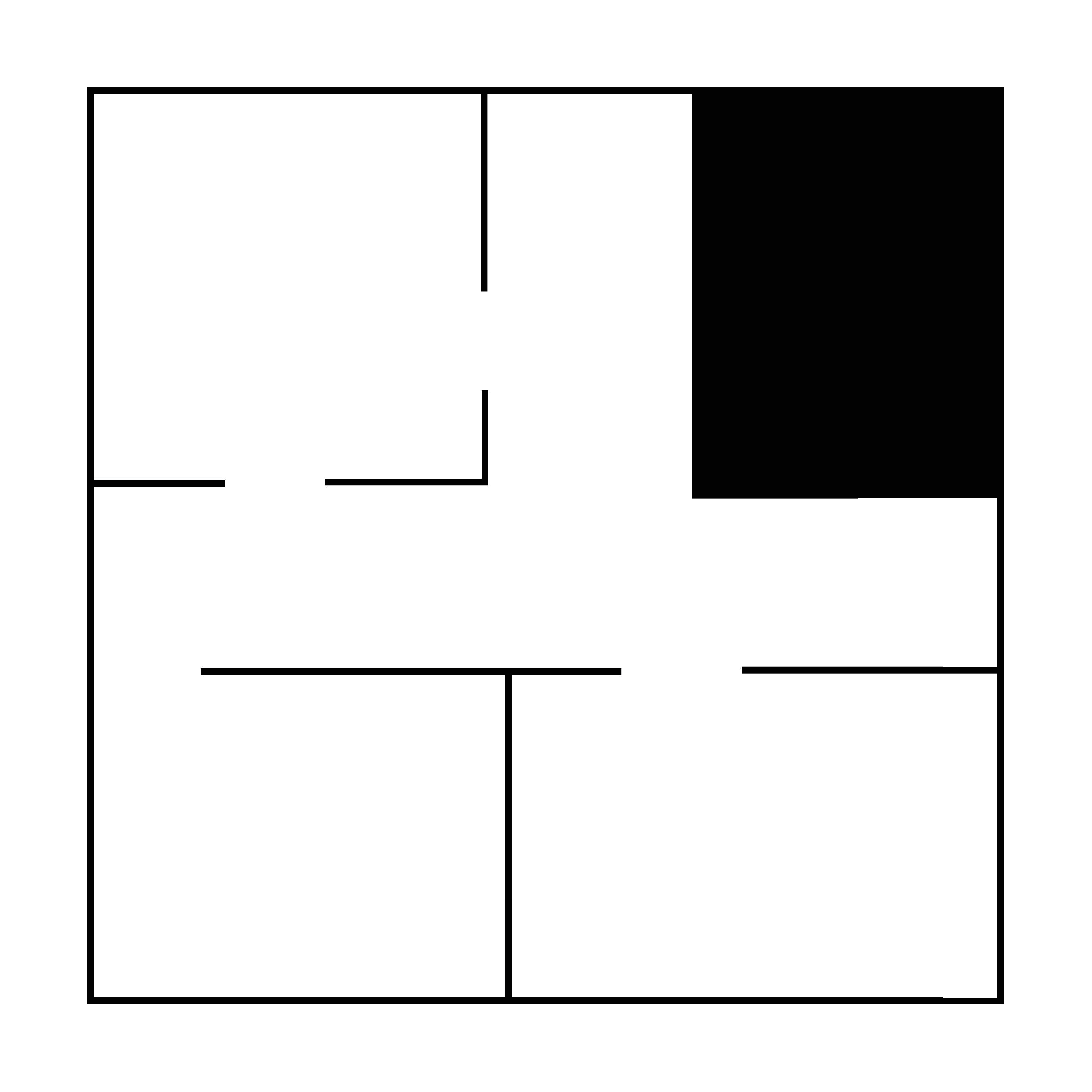}
        }\hspace{-1.2em}
        \subfloat{
            \includegraphics[width=0.165\linewidth, height=0.165\linewidth]{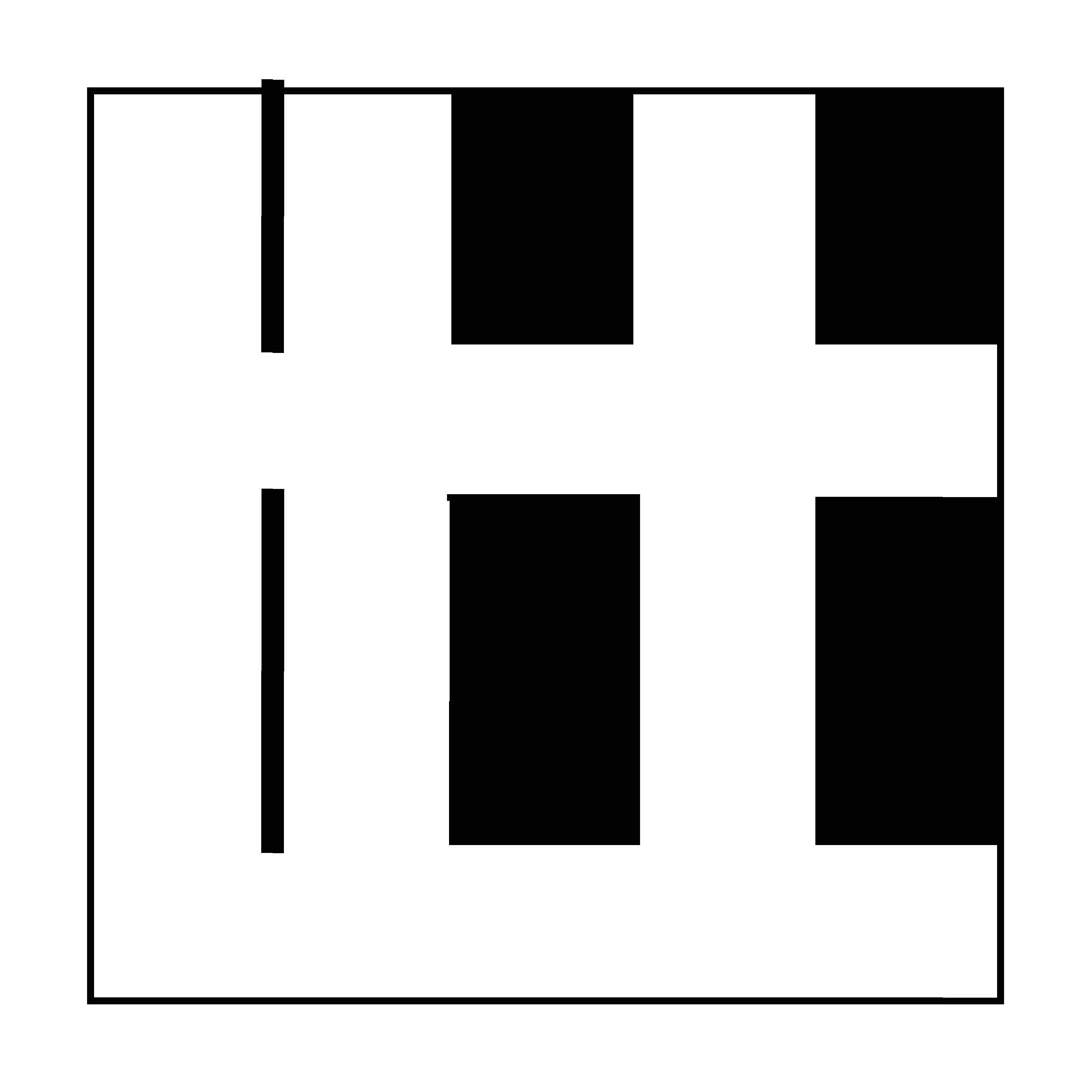}
        }\hspace{-1.2em}
        \subfloat{
            \includegraphics[width=0.165\linewidth, height=0.165\linewidth]{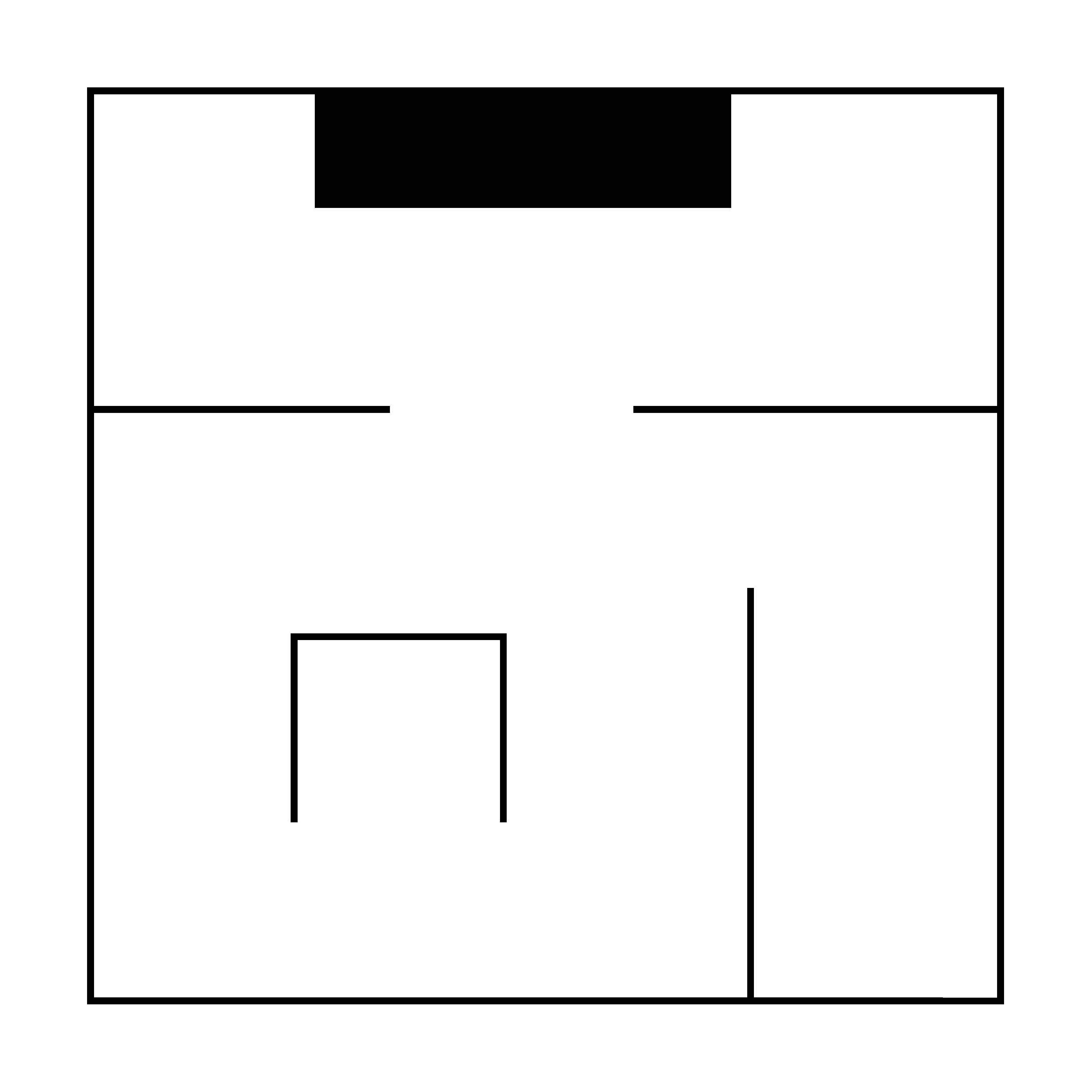}
        }
    \end{minipage}

    \vspace{0cm} % 添加垂直间距

    % 第二行子图
    \begin{minipage}{\linewidth}
        \centering
        \subfloat{
            \includegraphics[width=0.165\linewidth, height=0.165\linewidth]{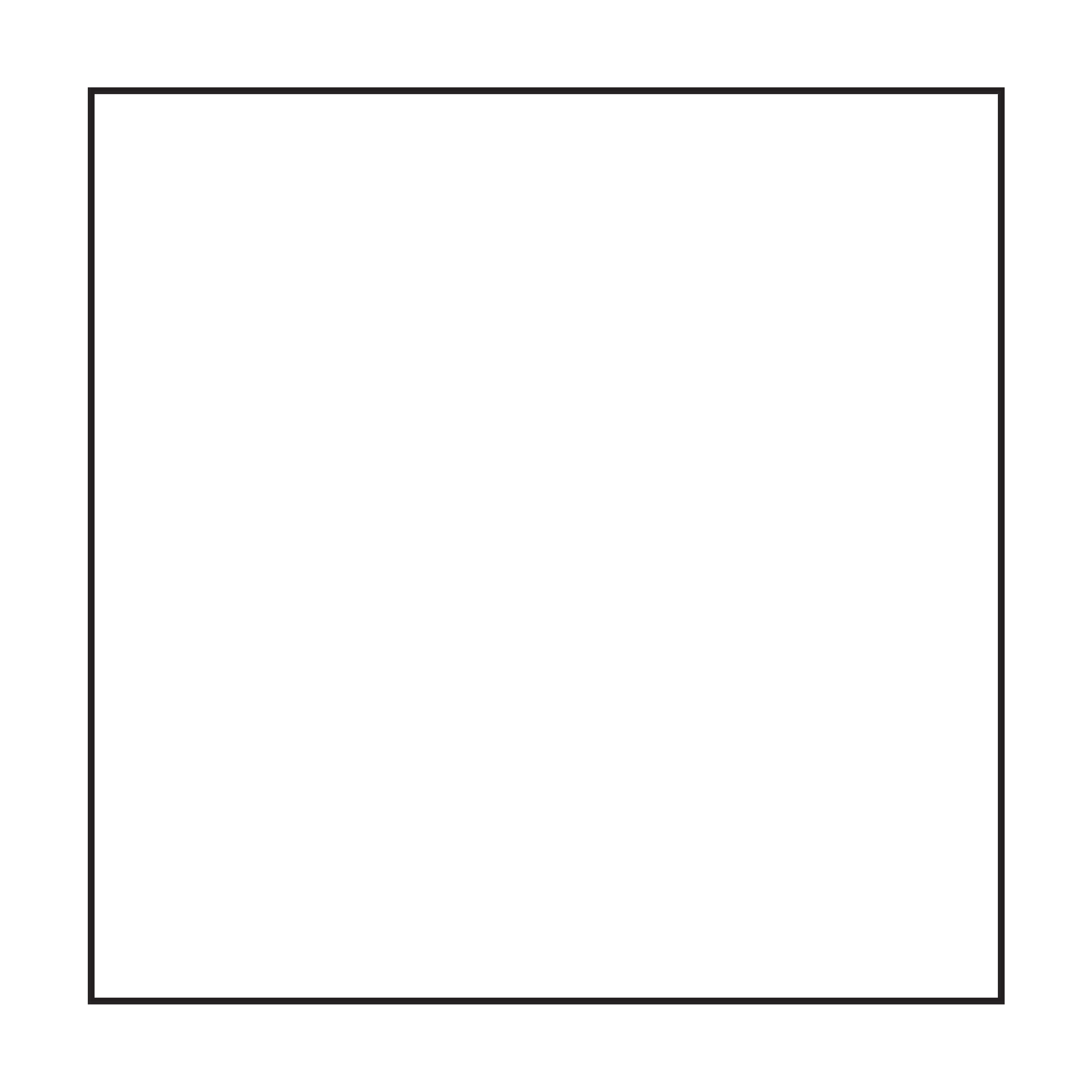}
        }\hspace{-1.2em}
        \subfloat{
            \includegraphics[width=0.165\linewidth, height=0.165\linewidth]{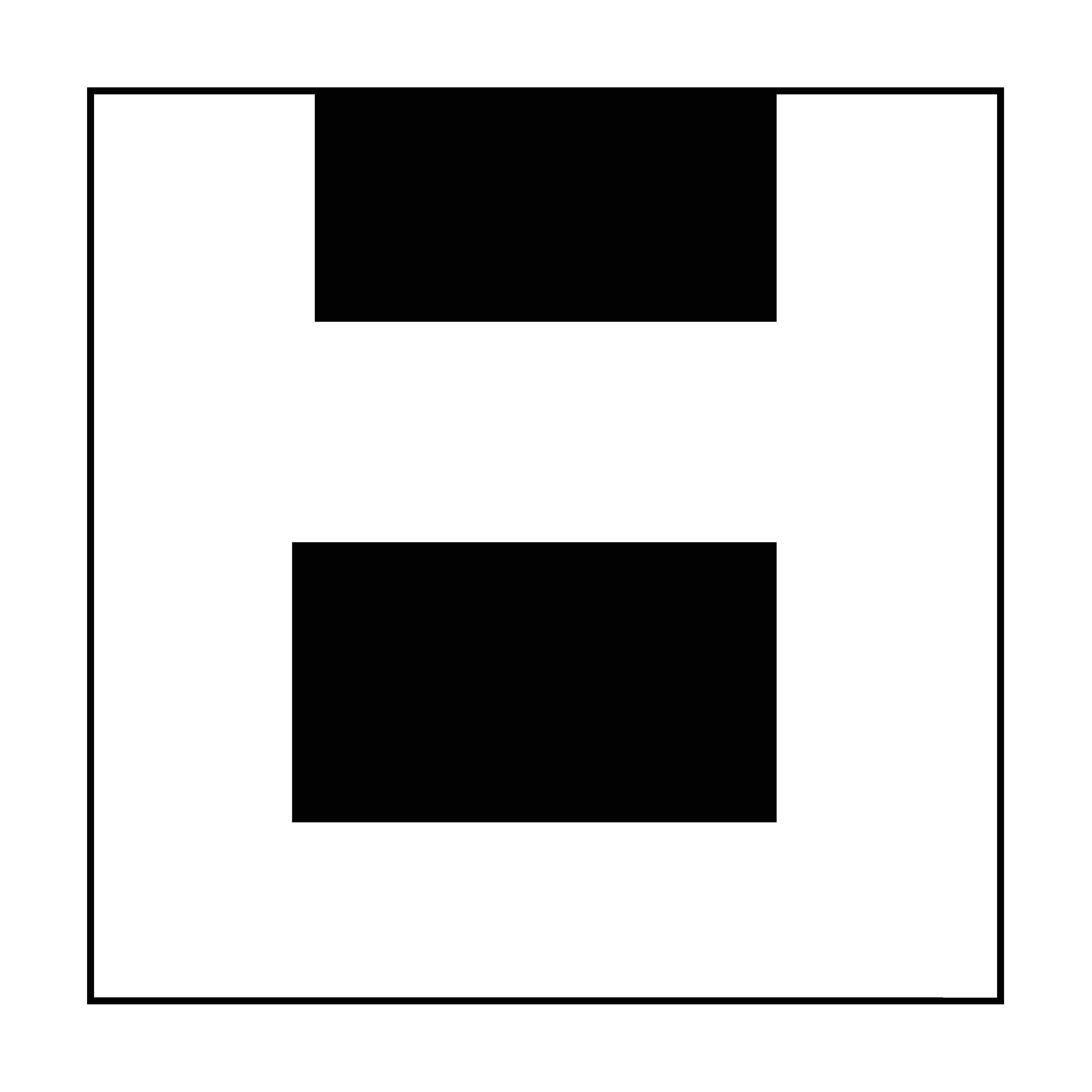}
        }\hspace{-1.2em}
        \subfloat{
            \includegraphics[width=0.165\linewidth, height=0.165\linewidth]{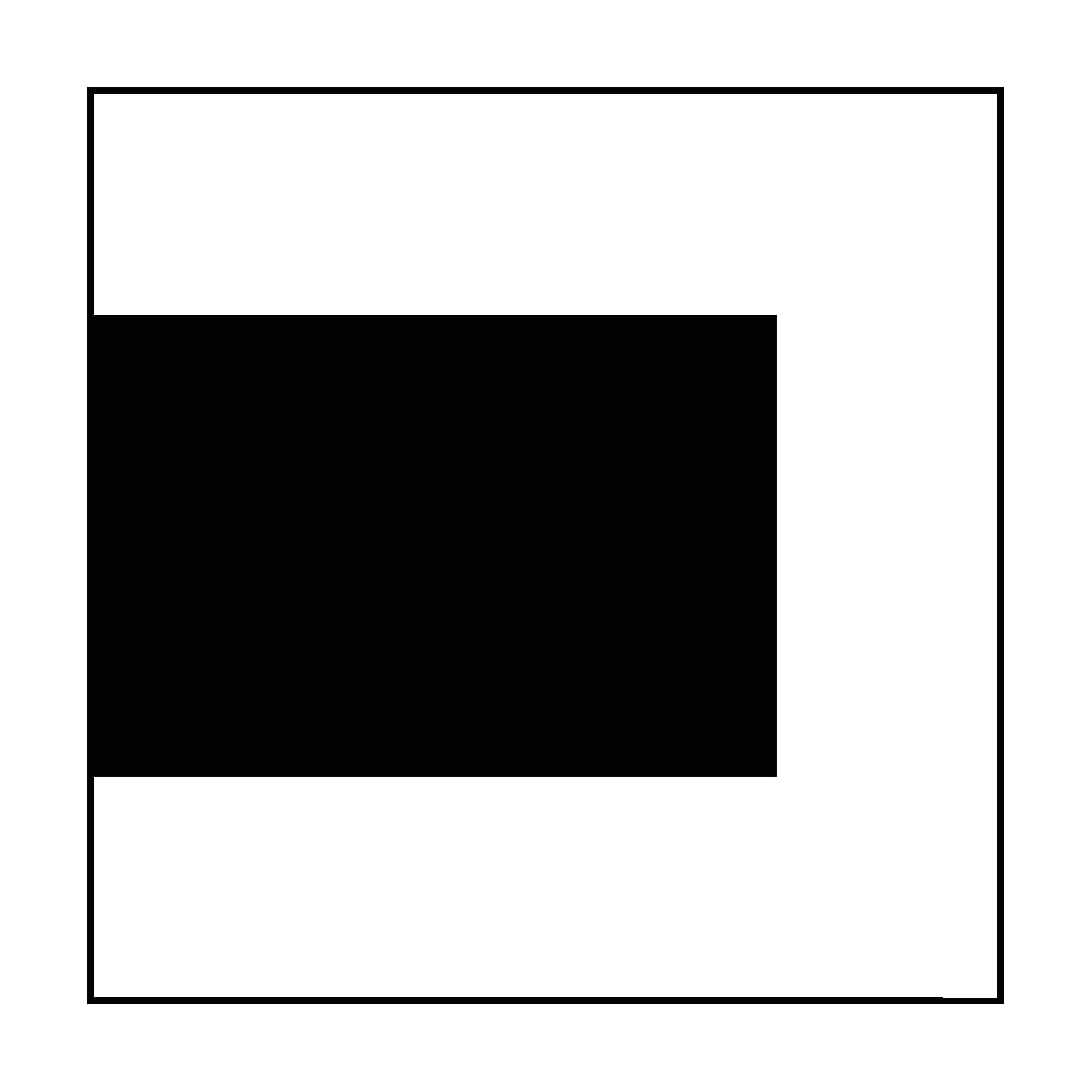}
        }\hspace{-1.2em}
        \subfloat{
            \includegraphics[width=0.165\linewidth, height=0.165\linewidth]{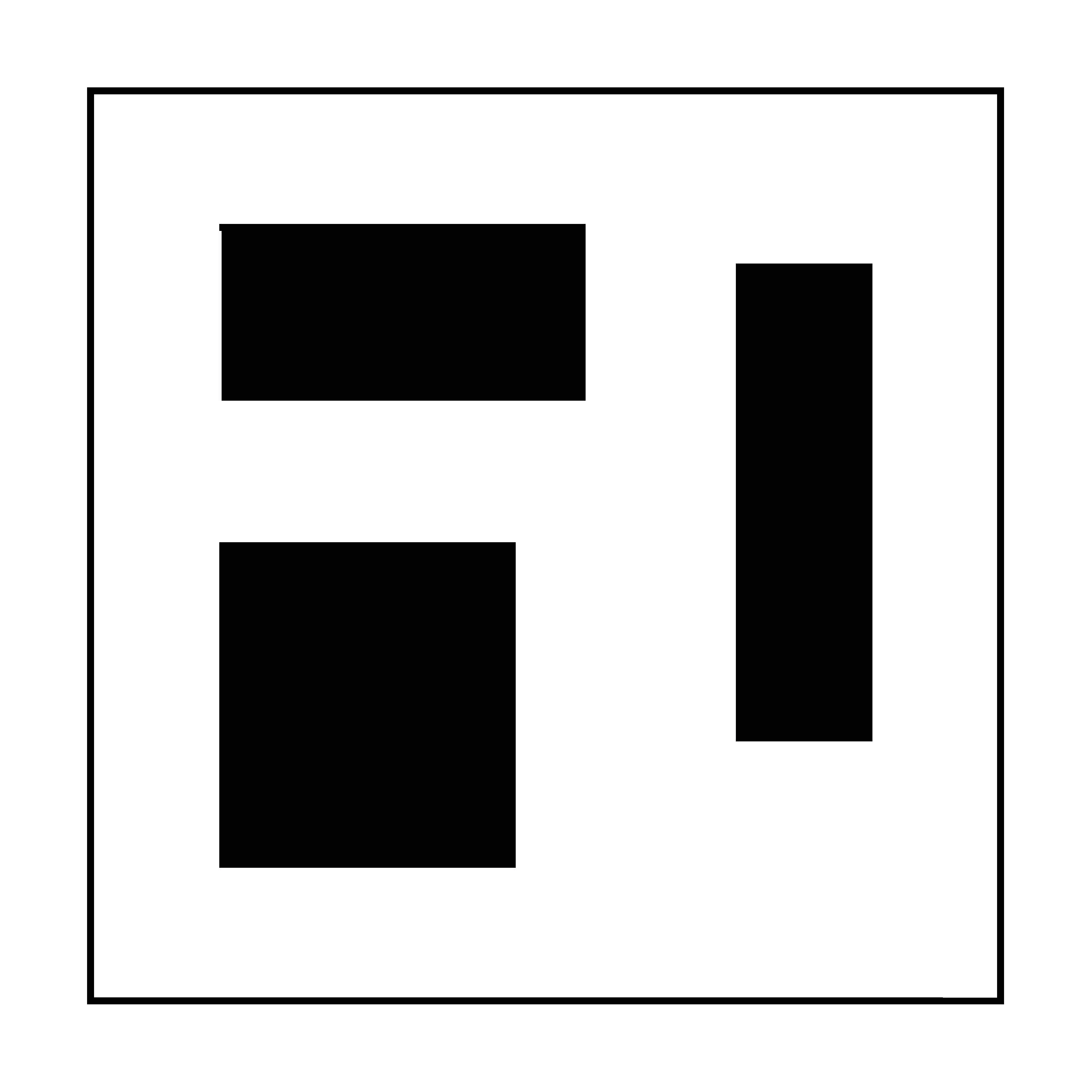}
        }\hspace{-1.2em}
        \subfloat{
            \includegraphics[width=0.165\linewidth, height=0.165\linewidth]{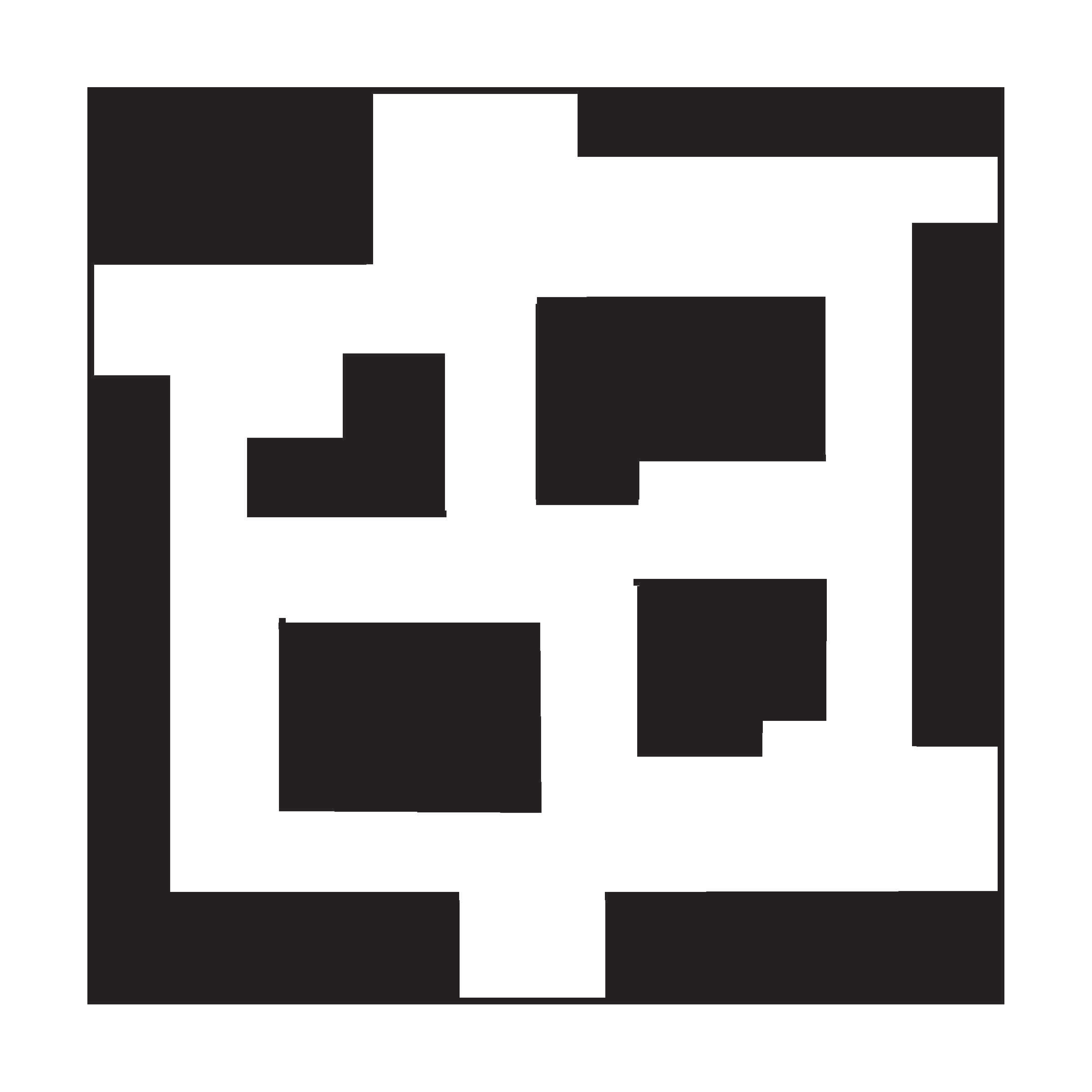}
        }\hspace{-1.2em}
        \subfloat{
            \includegraphics[width=0.165\linewidth, height=0.165\linewidth]{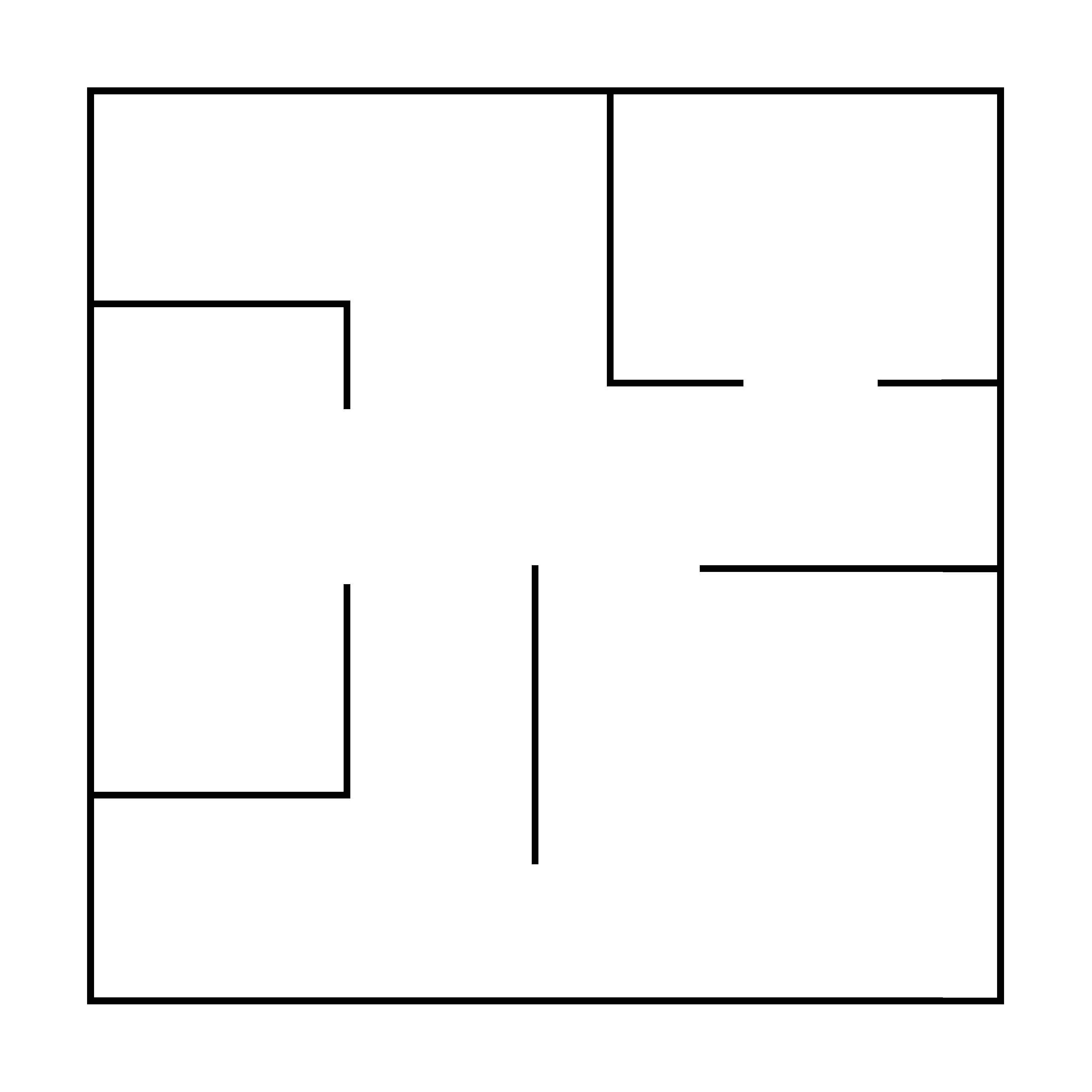}
        }
    \end{minipage}

    \caption{\textbf{Environment for different levels of training and testing.} The world in the first row are training worlds, the second row are testing worlds, and the graphic shows a 2D schematic.}
    \label{FigCombined}
\end{figure}

\begin{algorithm}[htbp]
    \caption{Periodic review-based\\curriculum learning for TSAC}
    \resizebox{0.9\linewidth}{!}{ % 调整宽度为90%
    \begin{minipage}{\linewidth}
    \begin{algorithmic}[1]
        \STATE Initialize environment $E$, initialize $TSAC$, $j \leftarrow 0$, $i \leftarrow 0$, environment container ($EC$) $\leftarrow \emptyset$, \\$e_{(n)}$ represents the $j$-th level training environment.
        \FOR{$j = 1$ to $n$}
            \STATE $EC \leftarrow EC \cup \{ e_{(j)} \}$ 
            \STATE $success\_rate \leftarrow 0$
            \WHILE{$success\_rate \le \beta$}
                \STATE $E \leftarrow$ sample from $EC$ with probability $P_{(i,j)}$
                \STATE Perform $TSAC$ actions in environment $E$
                \STATE $success\_rate \leftarrow \frac{1}{n} \sum\limits_{k=1}^n$ success history $[k]$
            \ENDWHILE
        \ENDFOR
    \end{algorithmic}
    \end{minipage}
    }
\end{algorithm}

Different levels of training environments are illustrated as shown in Fig. 6. The environments are designed in Gazebo, including dead ends, various sizes of obstacles, randomly appearing dynamic obstacles and situations where certain paths may be blocked due to random obstacles. These environments essentially cover typical scenarios encountered in AE. Each environment is modeled as an individual model, which can be generated using the spawn model feature and removed using the delete model function.

The pseudocode for curriculum learning is shown in Algorithm 1. In this study, the curriculum learning consists of six stages. With each stage advancement, the environment container adds worlds with corresponding difficulty. The generation probability of the $i$-th environment at stage $j$ is given by (9):
\begin{equation}
    p_{(i,j)}=\frac{i^2}{\sum_{k=1}^jk^2}
\end{equation}

\section{EXPERIMENTS AND DISCUSSION}

The proposed CTSAC was compared with state-of-the-art non-learning methods: Far Planner (FP) \cite{c8} and Rapidly-Exploring Random Tree (RRT*) \cite{c36} combined with Dynamic Window Approach (DWA) for navigation and learning-based methods: TD3 \cite{c31}. The parameters for the baseline algorithms were set to their default recommended values from open-source implementations.

\subsection{Simulation experiment}
In the simulation experiment, test worlds were designed as shown in the second row of Fig. 6. The platform used Ubuntu 20.04 with Intel i5 12400 and NVIDIA RTX 4080 GPU. The simulation environment uses Gazebo with a scene size of $20 \mathrm{m} \times 20 \mathrm{m}$. The robot used is a TurtleBot3, with a linear velocity range of $[0,\,1]\,\text{m/s}$ and an angular velocity range of $[-1,\,1]\,\text{rad/s}$. A Velodyne VLP16 LiDAR is used, with a detection range of $6 \, \mathrm{m}$ and a $360^\circ$ scanning field.

To evaluate generalization, six test worlds not included in the training dataset were used. Each robot was allowed a maximum of 1000 iterations, with 100 runs per world, and tested under the same initial and goal positions. The Success Rate (SR) and Success Rate Weighted Exploration Time (SET) for four algorithms were recorded.

As shown in Fig. 7(a), CTSAC quickly reached the goal, benefiting from Transformer-based SAC decision-making. The TD3 algorithm struggled to find the narrow entrance due to limited detection accuracy in the front, while the FP algorithm got stuck in a dead end due to a lack of environmental understanding. The RRT* algorithm followed walls, posing safety concerns, but CTSAC avoided this behavior. In Fig. 7(d), TD3 performed best in World 4 but lacked generalization. CTSAC achieved high success rates and shorter exploration times across all environments, indicating optimal performance. Learning-based methods, despite higher variance, demonstrate greater flexibility and generalization.

\begin{figure}[!h]
    \vskip 0.18cm
    \centering
    \setcounter{subfigure}{0}

    % 第一行子图
    \subfloat[World 6-1]{
        \begin{minipage}[t]{0.3\linewidth}
            \centering
            \includegraphics[height=0.99in]{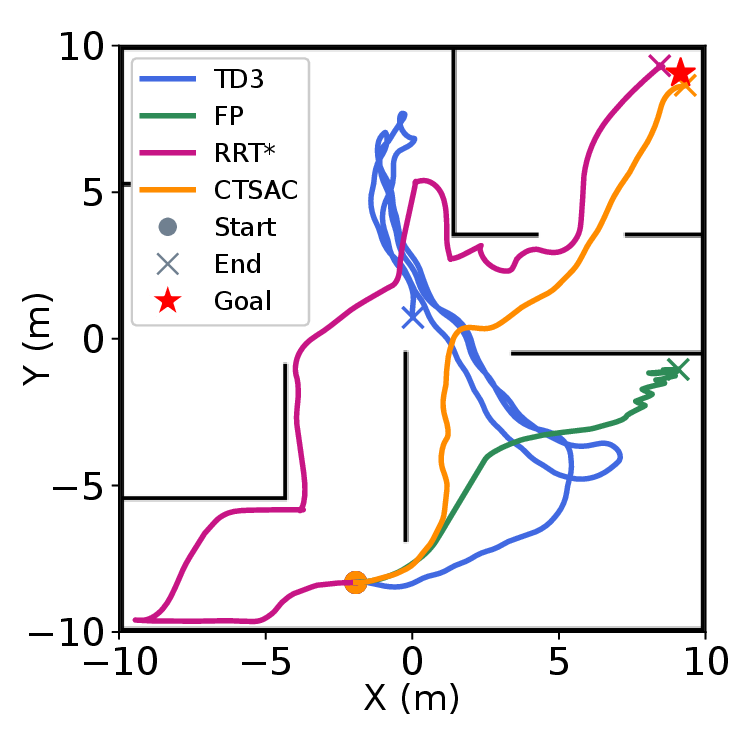}
        \end{minipage}% 
    }
    \hspace{-0.5em}
    \subfloat[World 6-2]{
        \begin{minipage}[t]{0.3\linewidth}
            \centering
            \includegraphics[height=0.99in]{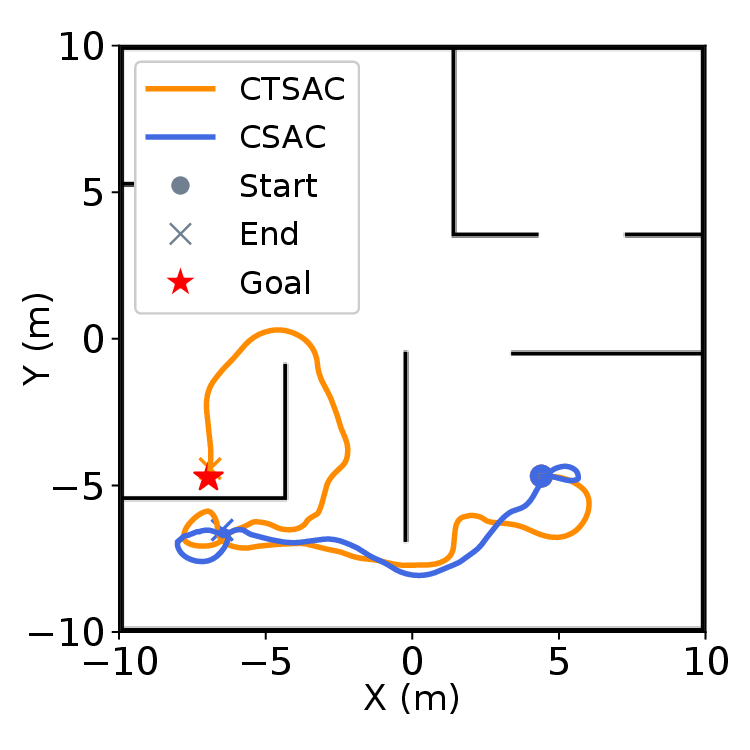}
        \end{minipage}% 
    }
    \hspace{-0.5em}
    \subfloat[World 6-3]{
        \begin{minipage}[t]{0.3\linewidth}
            \centering
            \includegraphics[height=0.99in]{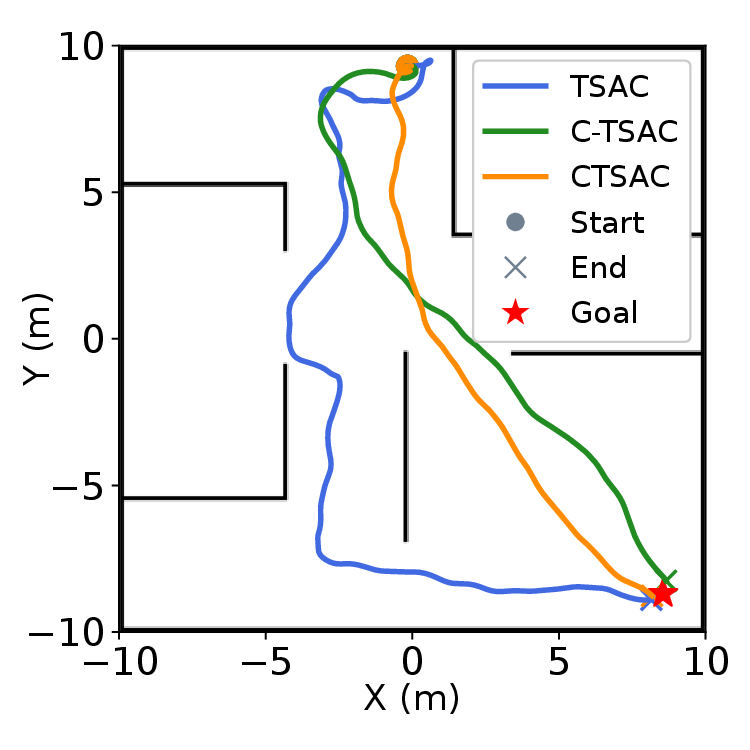}
        \end{minipage}% 
    }

    \vspace{0cm} % 添加垂直间距

    % 第二行子图的第一行
    \captionsetup[subfigure]{labelformat=empty}
    \subfloat[]{
        \begin{minipage}[t]{0.3\linewidth}
            \centering
            \includegraphics[height=0.7in]{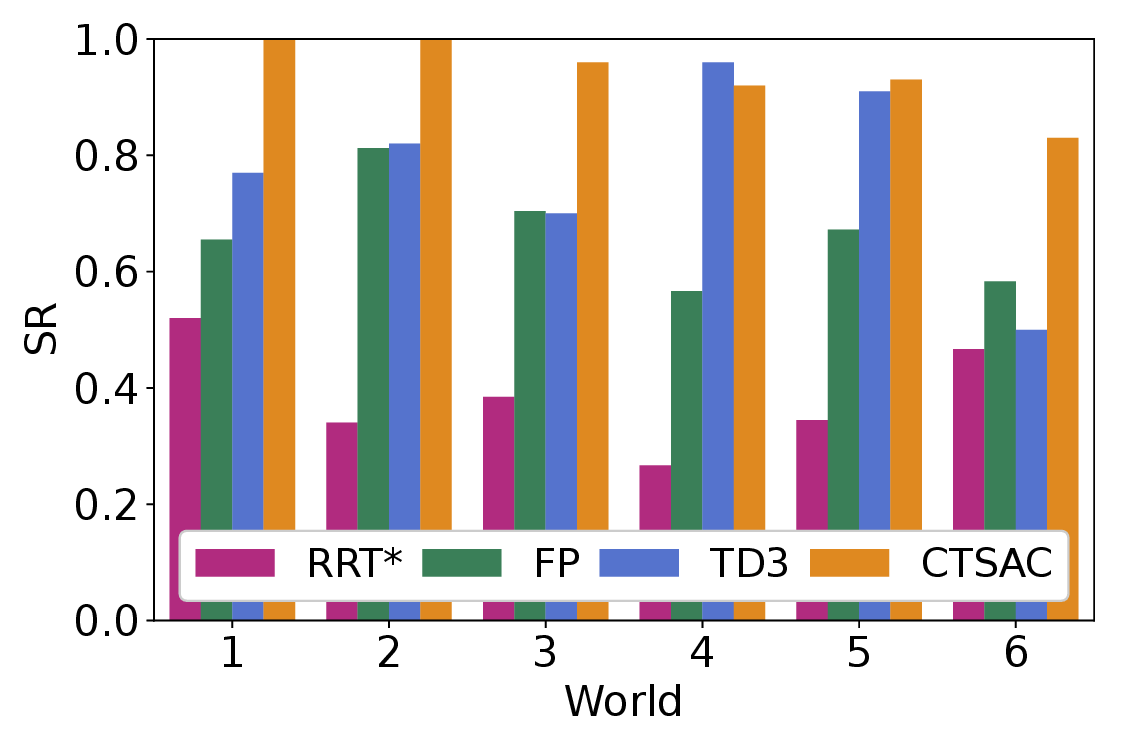}
        \end{minipage}% 
    }
    \hspace{-0.5em}
    \subfloat[]{
        \begin{minipage}[t]{0.3\linewidth}
            \centering
            \includegraphics[height=0.7in]{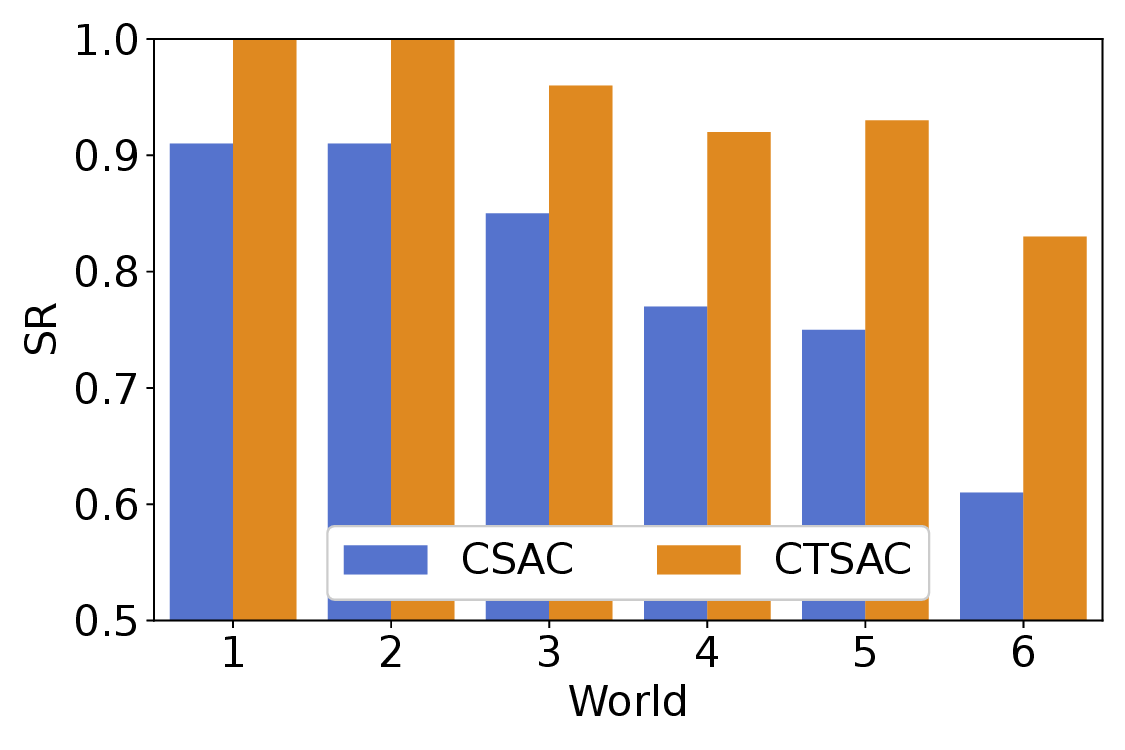}
        \end{minipage}% 
    }
    \hspace{-0.5em}
    \subfloat[]{
        \begin{minipage}[t]{0.3\linewidth}
            \centering
            \includegraphics[height=0.7in]{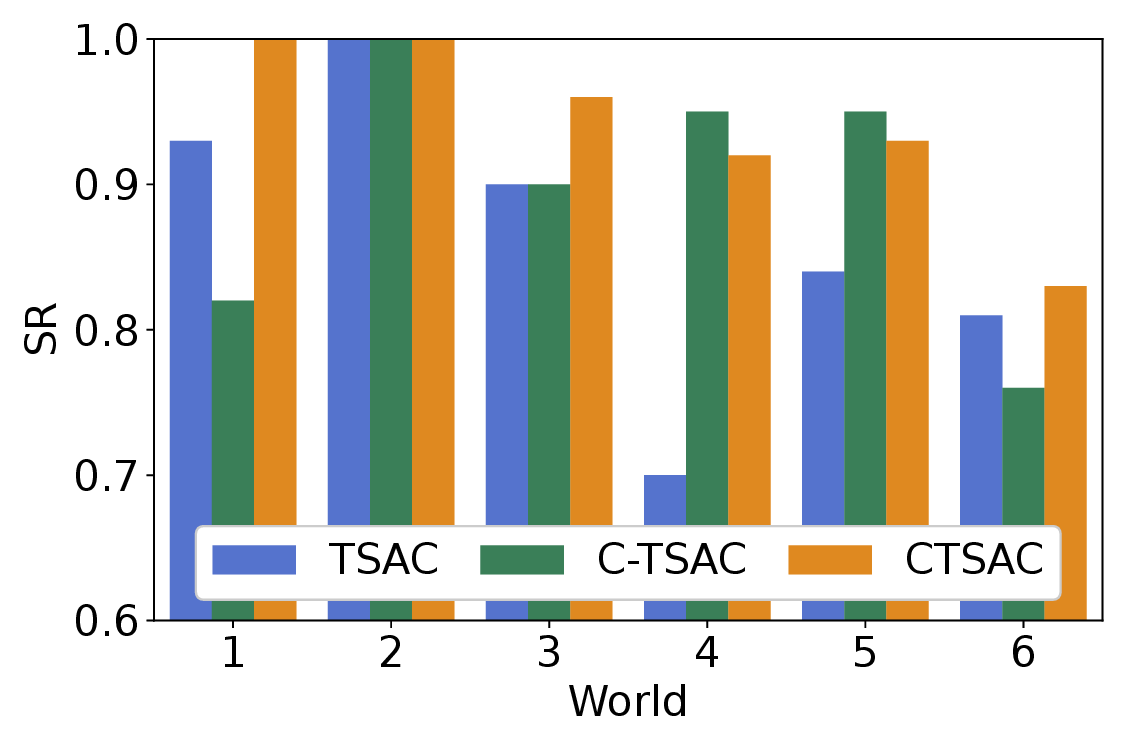}
        \end{minipage}% 
    }
    \captionsetup[subfigure]{labelformat=parens} % 恢复默认设置

    \vspace{-0.6cm} % 添加垂直间距

    % 第二行子图的第二行
    \setcounter{subfigure}{3} % 设置子图计数器为3，从d开始
    \subfloat[PR 1]{
        \begin{minipage}[t]{0.3\linewidth}
            \centering
            \includegraphics[height=0.7in]{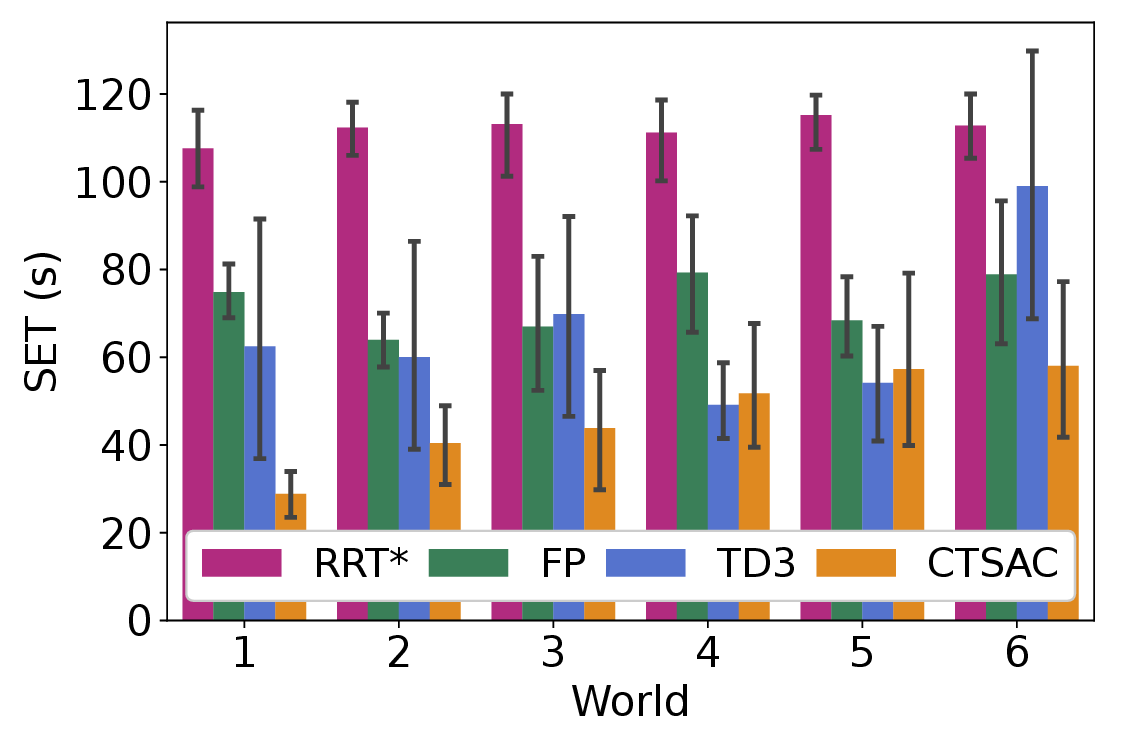}
        \end{minipage}% 
    }
    \hspace{-0.5em}
    \subfloat[PR 2]{
        \begin{minipage}[t]{0.3\linewidth}
            \centering
            \includegraphics[height=0.7in]{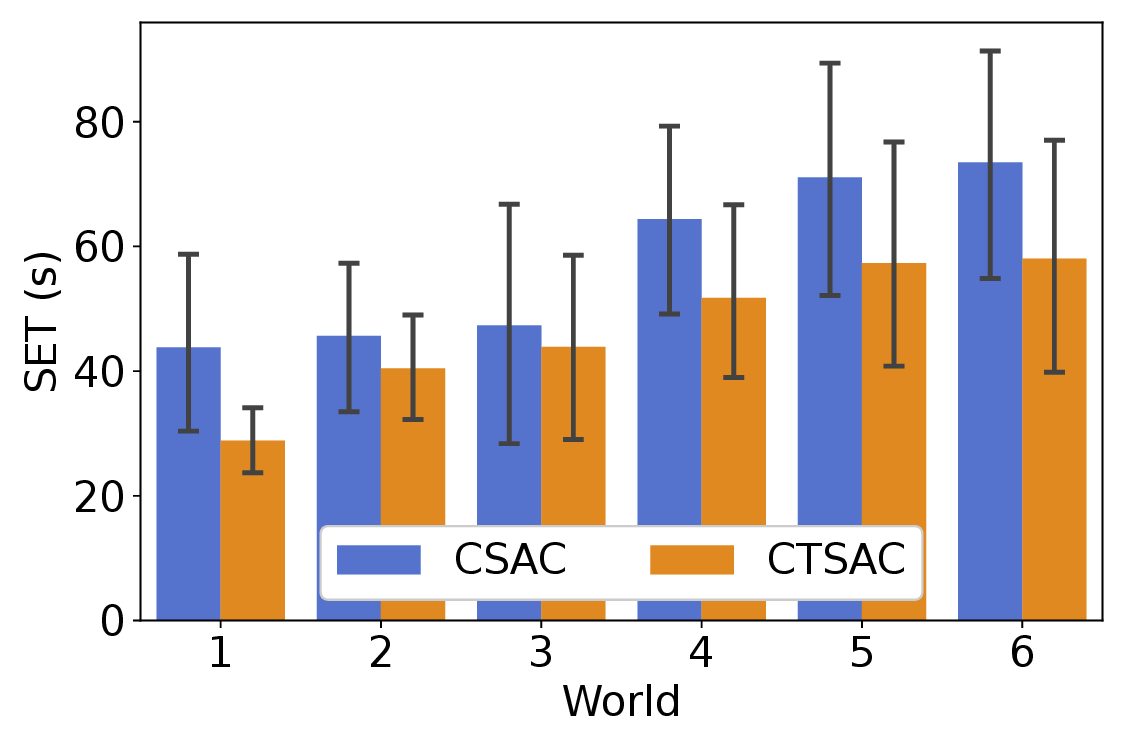}
        \end{minipage}% 
    }
    \hspace{-0.5em}
    \subfloat[PR 3]{
        \begin{minipage}[t]{0.3\linewidth}
            \centering
            \includegraphics[height=0.7in]{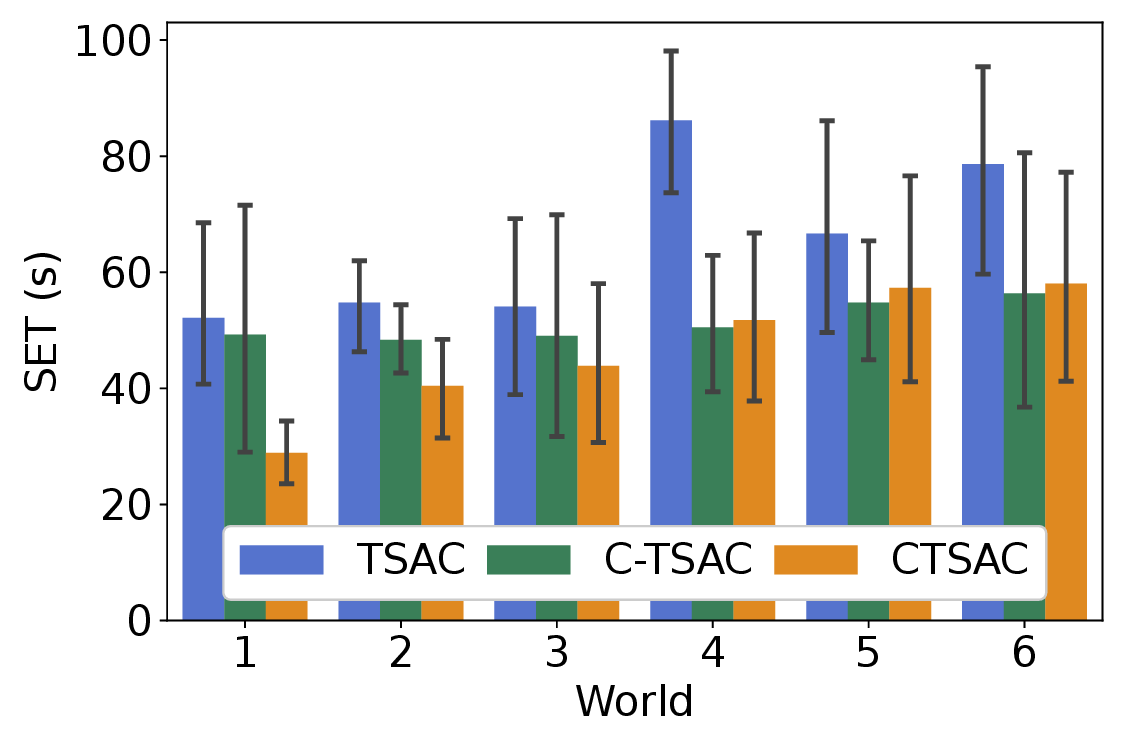}
        \end{minipage}% 
    }

    \vspace{0cm} % 添加垂直间距

    % 第三行子图
    \begin{minipage}[c]{0.45\linewidth}  % Use vertical center alignment
        \centering
        \includegraphics[height=0.8in]{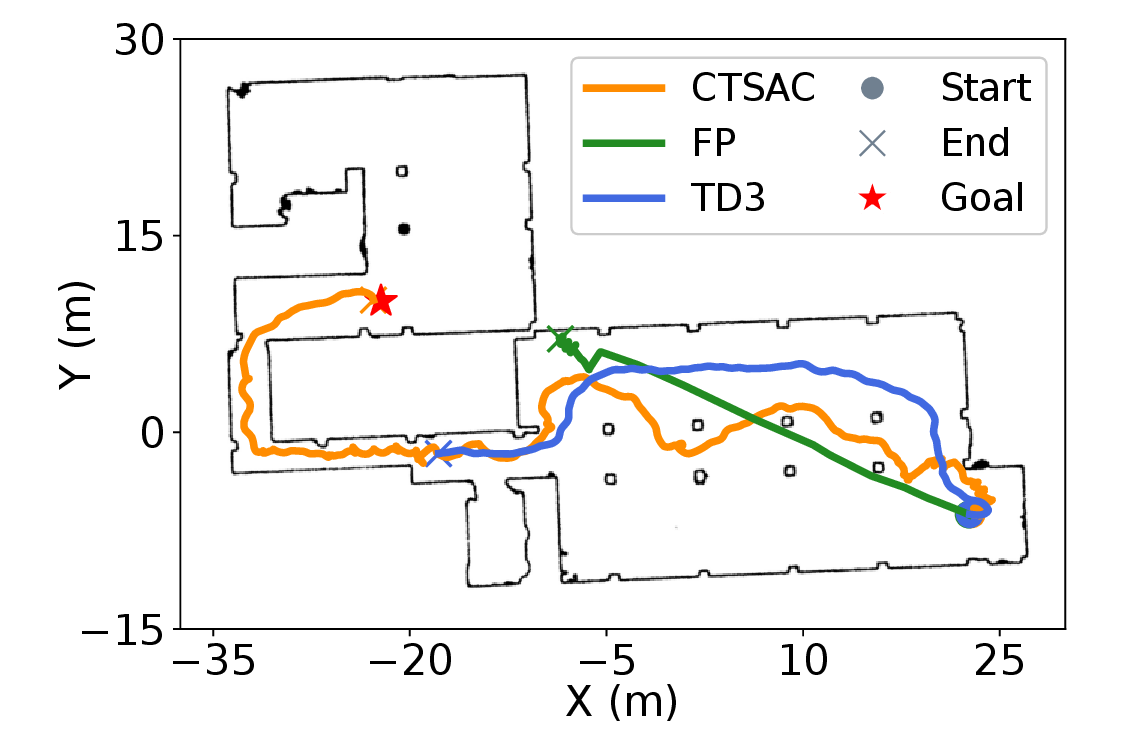}
        \subcaption{Exploration trajectory}
    \end{minipage}%
    \hspace{1em}  % Space between the two figures
    \begin{minipage}[c]{0.45\linewidth}  % Use vertical center alignment for the second figure
        \centering
        \includegraphics[height=0.8in]{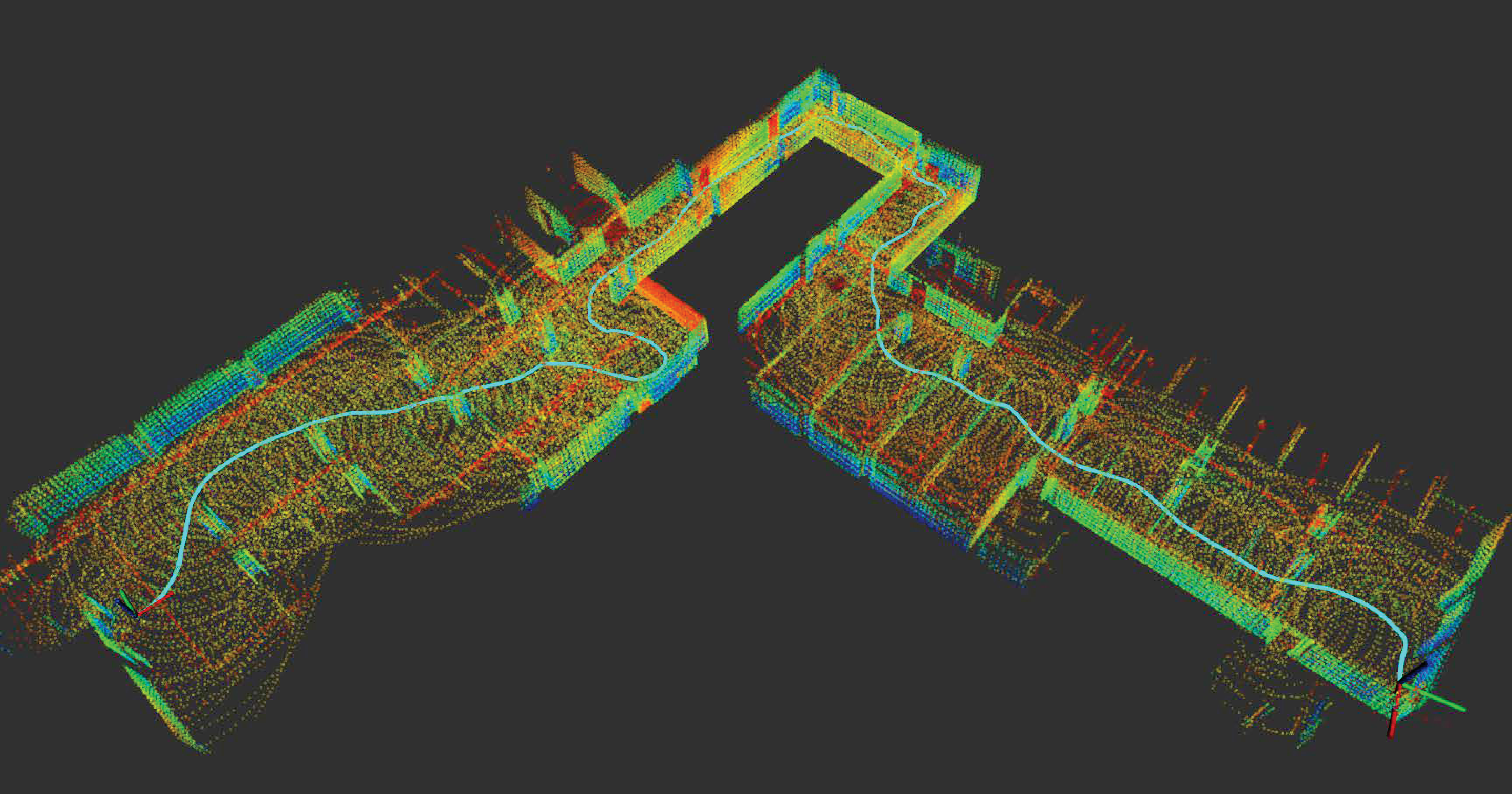}  % Adjusted height to match
        \subcaption{Global map}
    \end{minipage}

    \caption{\textbf{Experimental trajectory diagram, SR, and SET}, (a)-(c) show trajectory plots in simulation environment world5. (d)-(f) show Performance Ssults (PR) plots in various test worlds, including SR and SET. Histogram means the average value for different metrics and the error bar is the range under a 95\% confidence interval. (g) shows trajectory plots of real-world experiments, and (h) shows the map built during exploration.}
    \label{fig7}
\end{figure}

\subsection{Ablation experiment}
\subsubsection{Transformer}
To validate that the introduction of the Transformer structure can enhance decision-making quality and effectively address the issues of robots wandering in place and escaping local optima, we compared CTSAC with CSAC (the SAC algorithm without the Transformer). Both models were trained under identical settings.

As shown in Fig. 7(b), in an environment with the same dead end, CTSAC successfully navigated past the obstacle, while CSAC continued to wander. CTSAC effectively utilizes the self-attention mechanism of the Transformer to capture long-term historical information, enabling it to make more accurate decisions and avoid stagnation. Furthermore, as depicted in Fig. 7(e), the SR and SET of CTSAC outperform those of CSAC by 10\%. This demonstrates that CTSAC not only improves task success rates but also optimizes exploration efficiency by reducing redundant behaviors. The performance gap becomes even more pronounced in more complex environments.

\subsubsection{Curriculum learning}
A comparative experiment on periodic review-based curriculum learning was conducted to evaluate the generalization performance of CTSAC and its ability to mitigate the forgetting effect. The experiment compared CTSAC, C-TSAC (traditional switching-based curriculum) and TSAC (without curriculum learning) in terms of generalization performance on test worlds. All three algorithms, based on the Transformer-based SAC, were trained under identical settings.

As shown in Fig. 7(c)(f), TSAC exhibited poor generalization, as it was trained only on the final world, limiting its acquired knowledge. This highlights the importance of curriculum learning in improving the generalization performance of RL. C-TSAC showed moderate generalization by learning from diverse worlds, but still experienced some forgetting in the simpler world 1. CTSAC, on the other hand, maintained a stable success rate across all environments, being 20\% higher than C-TSAC in world 1, with minimal performance gap in more complex environments. This suggests that CTSAC, through its periodic review-based curriculum learning mechanism, effectively mitigated the forgetting effect and showed enhanced generalization performance across a variety of environments.

\subsection{Real-world experiment}
\begin{table}[!h]
    \centering
    \resizebox{0.7\linewidth}{!}{
        \begin{tabular}{c|c c c c }
            \hline
            & SR & \multicolumn{2}{c}{SET(s)} \\ 
            \hline
            FP   & 0.575 & $104.83 \pm 30.59$  &\\ 
            TD3  & 0.775 & $59.98 \pm 20.62$  &\\ 
            CTSAC& \textbf{0.8} & $\bm{25.79 \pm 18.59}$ &\\ 
            \hline
        \end{tabular}
    }
    \caption{Performance results from real-world experiments.}
    \label{tab:addlabel}
\end{table}

To evaluate the S2R performance of CTSAC in real-world settings, as shown in Fig. 1, an underground parking lot covering an area of 45 by 60 $\mathrm{m}$ was selected as the test site. The environment features pillars, narrow passages, doors that may open or close, and pedestrians. CTSAC was tested and compared with FP and TD3, with multiple starting and goal points set. The robot used for testing was an AgileX Bunker, equipped with an NVIDIA Jetson Orin NX and a Velodyne VLP16 LiDAR. Each algorithm was tested 40 times.

As shown in Table 1, CTSAC achieved the highest success rate of 0.8, which is 22\% higher than FP and comparable to TD3. However, CTSAC demonstrated a shorter exploration time, indicating more efficient path planning. It also exhibited good transferability from simulation to real-world environments. In Fig. 7(g), CTSAC successfully escaped the local optimum where FP got stuck, navigating through the narrow passage to reach the goal. In contrast, TD3 failed to pass through the narrow passage due to its poor observation performance. We observed that the robot tended to slip, indicating that the velocity tracking controller did not effectively track the commands, which needs to be addressed in future work.

\section{CONCLUSION}

In this work, we propose CTSAC, a Transformer-based reinforcement learning method with curriculum learning for goal-oriented autonomous robot exploration. By integrating the Transformer into SAC, the robot is able to leverage historical information, enhancing its reasoning ability regarding the environment and improving the quality of its decision-making. Additionally, we introduced a periodic review-based curriculum learning approach that optimized the switching mechanism and mitigated catastrophic forgetting, thus improving the model's generalization performance. Furthermore, we conducted S2R transfer experiments, where CTSAC demonstrated excellent S2R performance, thanks to the exploration framework, continuous training environments, and LiDAR clustering optimization. Future work will draw inspiration from large-scale models and embodied intelligence approaches to further improve the model's performance in autonomous exploration.

\end{document}